\documentclass{article}

\usepackage{microtype}
\usepackage{graphicx}
\usepackage{subcaption}
\usepackage{booktabs} 

\usepackage{hyperref}

\usepackage[accepted]{icml2026}

\usepackage{amsmath}
\usepackage{amssymb}
\usepackage{mathtools}
\usepackage{amsthm}
\usepackage{multirow}

\usepackage[capitalize,noabbrev]{cleveref}

\theoremstyle{plain}

\theoremstyle{definition}

\theoremstyle{remark}

\usepackage[textsize=tiny]{todonotes}

\begin{document}

\twocolumn[
  \icmltitle{FG-CLIP 2: A Bilingual Fine-grained Vision-Language Alignment Model}



  \icmlsetsymbol{equal}{*}

    \begin{icmlauthorlist}
    \icmlauthor{Chunyu Xie}{yyy,equal}
    \icmlauthor{Bin Wang}{yyy,equal}
    \icmlauthor{Fanjing Kong}{yyy}
    \icmlauthor{Jincheng Li}{yyy}
    \icmlauthor{Dawei Liang}{yyy}
    \icmlauthor{Ji Ao}{yyy}
    
    \icmlauthor{Dawei Leng}{yyy}
    \icmlauthor{Yuhui Yin}{yyy}
    \end{icmlauthorlist}

    \begin{center} 
    Homepage: https://360cvgroup.github.io/FG-CLIP \\
    Code: https://github.com/360CVGroup/FG-CLIP\\ 
    Model\&Dataset: https://huggingface.co/collections/qihoo360/fg-clip-2
    \end{center}
    
    \icmlaffiliation{yyy}{360 AI Research}
    
    \icmlcorrespondingauthor{Dawei Leng}{lengdawei@360.cn}
    
    \icmlkeywords{Machine Learning, ICML}

  \vskip 0.3in
]

\def\bin{\textcolor{black}}
\def\xie{\textcolor{black}}
\def\iclrre{\textcolor{black}}
\newcommand{\fix}{\marginpar{FIX}}
\newcommand{\new}{\marginpar{NEW}}
\def\ljc{\textcolor{black}}
\def\jc{\textcolor{black}}
\def\ch{\textcolor{black}}

\def\bc{\textcolor{black}}
\def\TODO{\textcolor{black}}
\def\HM{\textcolor{blue}}
\def\green{\textcolor{black}}
\def\red{\textcolor{red}}
\def\blue{\textcolor{black}}
\def\ldw{\textcolor{black}}

\def\lightgray{\textcolor{gray}}
\def\final{\textcolor{black}}


\printAffiliationsAndNotice{}  

\begin{abstract}
Fine-grained vision-language understanding requires precise alignment between visual content and linguistic descriptions, a capability that remains limited in current models, particularly in non-English settings. While models like CLIP perform well on global alignment, they often struggle to capture fine-grained details in object attributes, spatial relations, and linguistic expressions, with limited support for bilingual comprehension. To address these challenges, we introduce FG-CLIP~2, a bilingual vision-language model designed to advance fine-grained alignment for both English and Chinese. Our approach leverages rich fine-grained supervision, including region-text matching and long-caption modeling, alongside multiple discriminative objectives. We further introduce the Textual Intra-modal Contrastive (TIC) loss to better distinguish semantically similar captions. 
\final{Trained on a carefully curated mixture of large-scale English and Chinese data, including a newly released 12M Chinese region-text dataset, FG-CLIP~2 achieves powerful bilingual performance.}
To enable rigorous evaluation, we present a new benchmark for Chinese multimodal understanding, featuring long-caption retrieval and bounding box classification. Extensive experiments on 29 datasets across 8 tasks show that FG-CLIP~2 outperforms existing methods, achieving state-of-the-art results in both languages. We release the model, code, and benchmark to facilitate future research on bilingual fine-grained vision-language alignment.
\end{abstract}    
\section{Introduction}
\label{sec:intro}

Vision-language alignment models~\citep{siglip2,chuang2025metaclip2} have undergone rapid evolution in recent years, driven by pioneering works such as CLIP~\citep{openaiclip},  which introduced large-scale contrastive pre-training on image-text pairs and demonstrated remarkable success in learning joint multimodal representations. These models excel at global alignment tasks such as zero-shot image classification and image-text retrieval, forming the foundation for a wide range of multimodal understanding systems~\citep{zhu2025internvl3,team2025gemma,li2025hunyuangamecrafthighdynamicinteractivegame,wu2025qwenimagetechnicalreport,vteam2025glm45vglm41vthinkingversatilemultimodal}. Their ability to align visual and linguistic concepts without explicit supervision has enabled strong generalization to diverse scenarios, including visual question answering~\citep{lu2025ovis25technicalreport,wang2025iaa}, image captioning~\citep{bai2025qwen2.5,li2025eagle}, and content-based retrieval~\citep{zhang2025long,wei2024bevclip}. 
\xie{However, their performance often degrades on fine-grained understanding tasks that require discriminating between similar object attributes, spatial configurations, or semantic distinctions. Such tasks demand precise alignment at both visual and linguistic levels: visually, they involve recognizing objects, attributes, and their spatial arrangements; linguistically, they require distinguishing between semantically similar expressions. This performance gap stems from reliance on coarse-grained image-text pairs during training, which encourages thematic alignment while failing to capture the fine-grained correspondences essential for robust visual grounding or attribute recognition.}

\iclrre{Several recent works have sought to address these limitations. Approaches such as FineCLIP~\citep{jingfineclip} and LongCLIP~\citep{zhang2025long} improve fine-grained understanding by incorporating region-level signals or supporting longer textual inputs. FG-CLIP~\citep{xie2025fg} further advances fine-grained discrimination through large-scale data curation and attribute-aware hard negative sampling. However, achieving robust discrimination between highly similar textual descriptions or subtle visual variations remains challenging, suggesting opportunities for more advanced training objectives. Moreover, prior fine-grained approaches are predominantly developed and evaluated in English, leaving Chinese–English bilingual scenarios relatively underexplored. Meanwhile, models designed for Chinese, such as Chinese-CLIP~\citep{yang2022chinese} and R2D2~\citep{xie2023ccmb}, primarily focus on coarse-grained alignment and lack fine-grained modeling capability. Therefore, extending fine-grained alignment to bilingual vision–language settings remains an important yet underexplored problem.}

To address these challenges, we propose FG-CLIP 2, a unified framework for bilingual fine-grained vision-language alignment. Our training strategy employs a two-stage paradigm to progressively refine model capabilities. \xie{In the first stage, we perform initial global alignment with both short and long textual descriptions to capture coarse and detailed semantic content at the early phase of training.} In the second stage, we incorporate fine-grained learning objectives that improve regional alignment, discriminative capability, and cross-modal ranking performance. To further refine the model’s ability to distinguish similar region-level descriptions, we propose the Textual Intra-modal Contrastive (TIC) loss, which learns from filtered hard negatives among high-similarity text pairs. FG-CLIP 2 is trained on large-scale, high-quality bilingual datasets with careful curation, enabling strong performance in both English and Chinese across diverse fine-grained vision-language tasks.

\final{We further contribute a large-scale Chinese region-text dataset, FineRegion-CN, containing 12 million images with fine-grained region descriptions to fill the gap of Chinese region-level training data. Additionally, we introduce a new benchmark suite to advance evaluation in Chinese multimodal understanding, featuring challenging tasks such as long caption image-text retrieval and bounding box classification in Chinese that go beyond conventional short-text retrieval and assess fine-grained comprehension more rigorously. Extensive experiments show that FG-CLIP 2 outperforms existing models on 29 datasets across 8 vision-language tasks in both Chinese and English, demonstrating powerful bilingual fine-grained vision-language alignment capability. To support future research and real-world deployment, our model, training data, code, and benchmark are made publicly available.}

\section{Related Work}
\label{sec:related_work}

Vision-language alignment models trained on large-scale data, such as CLIP~\citep{openaiclip}, EVA-CLIP~\citep{sun2023eva}, SigLIP~\citep{zhai2023sigmoid}, MetaCLIP~\citep{xu2023metaclip} and DFN~\citep{dfn} have demonstrated strong zero-shot capabilities but primarily focus on global semantic alignment and are often trained on English-only corpora. While these models serve as backbones for downstream tasks including multimodal reasoning~\citep{bai2025qwen2.5}, open-vocabulary detection~\citep{fu2025llmdet}, and segmentation~\citep{cho2024catseg,xsam}, \iclrre{they lack fine-grained and multilingual understanding, limiting their broader applicability.}

Recent efforts aim to improve localization capability and dense feature alignment through region-level supervision. Methods like AlphaCLIP~\citep{sun2024alpha}, \iclrre{UMG-CLIP~\citep{shi2024umg},} FineCLIP~\citep{jingfineclip}, and FG-CLIP~\citep{xie2025fg} leverage bounding boxes or masked regions to enhance local correspondence, while \iclrre{CLOC~\citep{chen2024contrastive}}, TIPS~\citep{tips}, and SigLIP 2~\citep{siglip2} introduce architectural or training enhancements for richer feature generation. On the multilingual front, Chinese-CLIP~\citep{yang2022chinese} and R2D2~\citep{xie2023ccmb} target Chinese understanding, and MetaCLIP 2~\citep{chuang2025metaclip2} scales multilingual data collection for broader language coverage. However, these works often treat fine-grained and multilingual understanding separately, and none explicitly optimize both in a unified framework. This gap inspires our work.

\begin{figure*}[!tbp]
  \centering   \includegraphics[width=0.96\linewidth]{{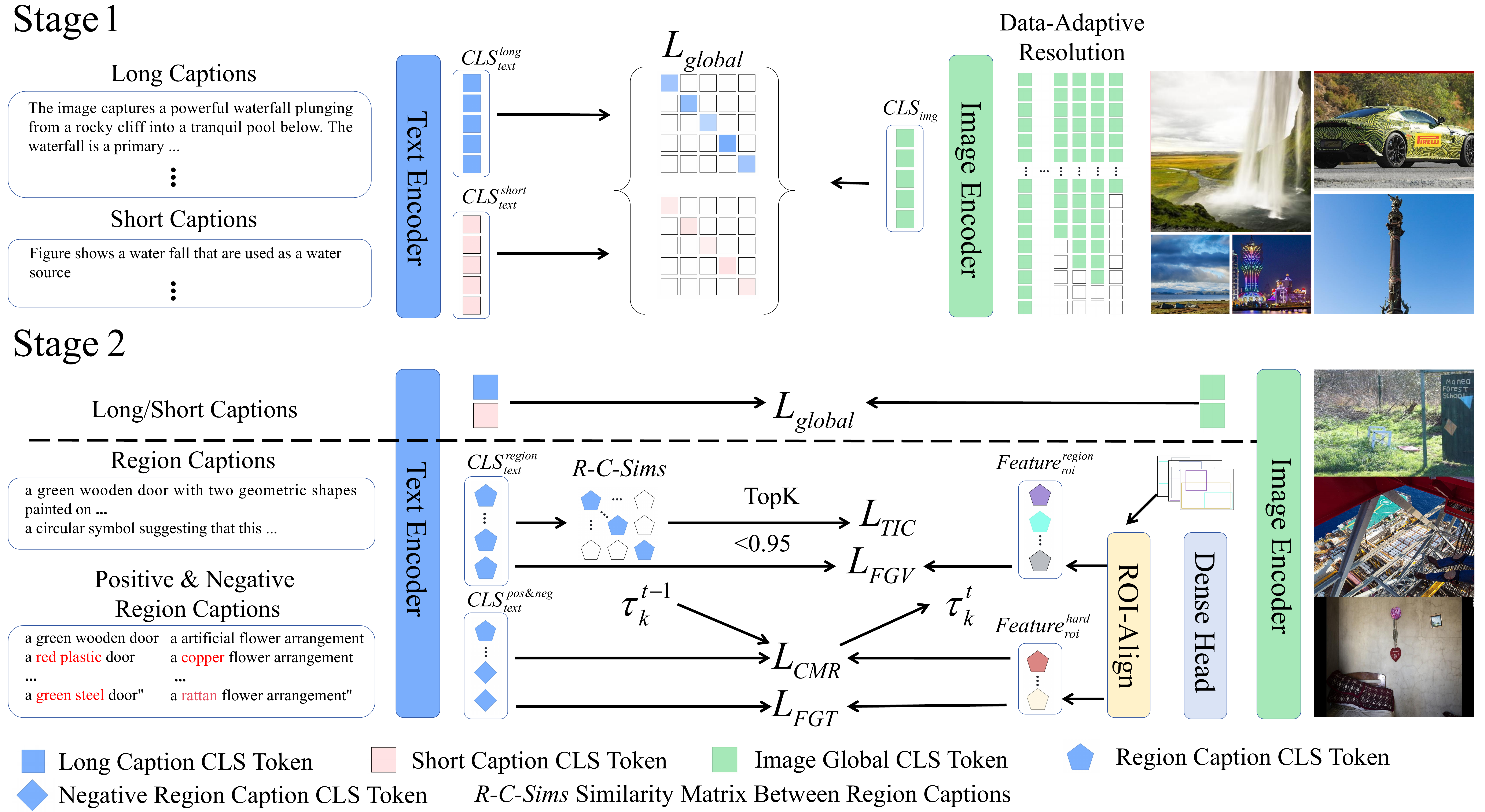}}
    \caption{\final{Overview of FG-CLIP 2. The framework consists of a two-stage training pipeline. Stage I performs bilingual global alignment using data-adaptive image resolution, matching image global tokens with long and short caption embeddings via \(\mathcal{L}_{\text{Global}}\). Stage II introduces fine-grained region-caption supervision alongside the global alignment. It employs ROI features and explicit hard-negative mining to compute auxiliary objectives: Fine-Grained Visual Learning (\(\mathcal{L}_{\text{FGV}}\)), Fine-Grained Textual Learning (\(\mathcal{L}_{\text{FGT}}\)), Textual Intra-modal Contrastive loss (\(\mathcal{L}_{\text{TIC}}\)), and Cross-modal Rank Loss (\(\mathcal{L}_{\text{CMR}}\)), thereby enhancing discrimination for semantic details.}}
    
    \label{fig:arch}
\end{figure*}

\section{Approach}
Figure~\ref{fig:arch} illustrates FG-CLIP 2, a vision-language model supporting Chinese and English. Our approach follows a two-stage hierarchical learning framework: the first stage establishes strong semantic alignment by training on large-scale image-text pairs, \xie{each associated with both a short caption and a long caption}; the second stage extends this learning by incorporating region-level alignment and fine-grained contrastive signals, enabling the model to preserve holistic scene understanding while enhancing its ability to discriminate fine-grained visual-language correspondences in both languages.

\subsection{Architecture}
We build upon the SigLIP 2~\citep{siglip2} dual-encoder framework, introducing key adaptations for fine-grained understanding and bilingual alignment. For the text encoder, we extend the maximum input length from 64 to 196 tokens to accommodate longer descriptions.
On the vision side, we adopt a data-adaptive resolution strategy: the target resolution is selected from \{128, 256, 576, 784, 1024\} based on the maximum image size per mini-batch, avoiding the stochastic sampling of SigLIP 2 and ensuring consistent training and inference behavior with minimal upscaling or downscaling. 
\iclrre{Global vision features are aggregated via a masked attention pooling head~\citep{zhai2022scaling}, while region embeddings are generated through an additional self-attention module. This design enables fine-grained, sub-image alignment and circumvents the information bottleneck caused by CLS-only pooling.}

\subsection{Training Objectives}

Our training proceeds in two stages. \iclrre{Stage I focuses on Global Alignment Learning. In Stage II, this objective is jointly optimized with four additional learning objectives, detailed in the subsequent paragraphs.}

\textbf{\xie{Global Alignment Learning}.}
We adopt the sigmoid loss from SigLIP~\citep{zhai2023sigmoid}, which treats image-text matching as a binary classification. For each image-text pair, similarity scores are computed across all pairs in the batch, and logistic regression is applied to distinguish positive from negative pairs. To enrich textual supervision, we include both original short captions and long captions generated by Large Multimodal Models (LMMs). This dual-caption strategy provides complementary signals: \xie{concise labels for global semantics and detailed descriptions for richer context.} \iclrre{We denote this global alignment loss as \(\mathcal{L}_{\text{Global}}\).}

\textbf{Fine-Grained Visual Learning.}
For each annotated region, we extract patch-level embeddings and apply RoIAlign~\citep{he2017mask} to obtain region-specific features. Corresponding text embeddings are derived from descriptions aligned to each bounding box. The regional contrastive loss encourages alignment between matched region-text pairs, promoting fine-grained cross-modal understanding. \iclrre{This loss term is denoted as \(\mathcal{L}_{\text{FGV}}\).}

\textbf{Fine-Grained Textual Learning.}
Following FG-CLIP~\citep{xie2025fg}, we leverage hard negatives from the FineHARD dataset. Each positive region-text pair is paired with 10 semantically similar negatives, constructed by perturbing key attributes (e.g., color, count, action) while preserving syntactic structure. 
\xie{The model is trained with a binary classification loss over the 1 positive and 10 negatives, encouraging the model to assign higher scores to the matched pair. This objective enhances the model’s ability to discriminate subtle textual differences.} \iclrre{We formulate this discriminative objective as \(\mathcal{L}_{\text{FGT}}\).}

\textbf{Textual Intra-modal Contrastive Loss.}
\xie{While cross-modal alignment ensures image-text correspondence, the text encoder itself often lacks sufficient discriminative pressure to separate semantically similar but distinct region descriptions, which is critical for fine-grained visual grounding. To address this, we propose the Textual Intra-modal Contrastive (TIC) loss, which operates purely within the text modality to sharpen the text encoder’s representation space.}
Given a batch of region texts, we compute pairwise similarities and filter out pairs with $\text{sim} > 0.95$ to avoid over-penalization. The top-10 most similar texts per sample are selected as hard negatives. The TIC loss is then defined as:
\begin{equation}
    \mathcal{L}_{\text{TIC}} = -\sum_{i=1}^{N} \log \frac{1}{\sum_{T_m \in \mathcal{T}_i} \exp(S(T_i, T_m))},
\end{equation}
\xie{where $\mathcal{T}_i$ denotes the set of filtered hard negatives for text $T_i$. This encourages the text encoder to assign lower similarity to hard negative pairs, thereby improving its ability to separate semantically close but distinct region descriptions.}

\textbf{Cross-modal Rank Loss with Global Threshold Synchronization.}
We adopt the Cross-modal Rank (CMR) loss~\citep{zhang2024contrasting} to enforce a margin between positive and hard negative pairs, \xie{thereby strengthening the model’s discrimination of semantic boundaries}. For a positive pair $(I, T)$ and hard negative $T_k$, the loss is defined as:
\begin{equation}
\mathcal{L}_{\text{CMR}} = \max\left(0, S(I, T_k) - S(I, T) + \tau_k\right),
\end{equation}
where $S(\cdot,\cdot)$ denotes cosine similarity. \xie{At training step $t$,} the margin $\tau_k$ is synchronized globally across all GPUs via all-reduce, \xie{where $\mathcal{B}_{\text{global}}$ denotes the union of all local batches across GPUs}:
\begin{equation}
\tau_k^t = \frac{1}{|\mathcal{B}_{\text{global}}|} \sum_{(I,T) \in \mathcal{B}_{\text{global}}} \left( S^{t-1}(I, T) - S^{t-1}(I, T_k) \right),
\end{equation}
\xie{with $S^{t-1}(\cdot,\cdot)$ denoting similarities computed at the previous step. This design ensures stable and consistent thresholding in distributed training.}

\iclrre{In Stage II, the training objective combines these terms as:
\begin{equation}
    \mathcal{L} = \mathcal{L}_{\text{Global}} + \lambda_1 \mathcal{L}_{\text{FGV}} + \lambda_2 \mathcal{L}_{\text{FGT}} + \lambda_3 \mathcal{L}_{\text{TIC}} + \lambda_4 \mathcal{L}_{\text{CMR}},
\end{equation}
with fixed weights $\lambda_1=0.1$, $\lambda_2=0.5$, $\lambda_3=0.1$, $\lambda_4=0.4$, \xie{chosen to ensure stable and effective multi-objective optimization}. The selection strategy for loss weights is detailed in Appendix~\ref{appendix:hyper}.}
\subsection{Training Data}
\xie{In the first stage, we train on image-text pairs from diverse sources, with a particular emphasis on enhancing semantic depth and linguistic coherence. For English, we adopt an enhanced version of the LAION-2B dataset~\citep{xie2025fg}, \iclrre{which augments} the original short captions with detailed long captions generated by LMMs. The original captions are often fragmented and contain keyword stacking or irrelevant noise, making them insufficient for training models to understand rich, compositional language. We retain original captions to preserve diversity of natural language expressions, while simultaneously training on their long-caption counterparts. This dual-caption strategy enables the model to learn from both concise, real-world descriptions and semantically dense, contextually coherent narratives. For Chinese, we combine three datasets: Wukong (100M pairs)~\citep{gu2022wukong}, Zero (250M pairs)~\citep{xie2023ccmb}, and a large-scale in-house dataset (500M pairs). }

In the second stage, we extend training with fine-grained region-text pairs to further improve spatial grounding. For English, we use the FineHARD dataset~\citep{xie2025fg}, which includes 12 million images, 40 million bounding boxes with fine-grained region descriptions, and 10 million hard negative samples. 
\final{For Chinese, we construct a complementary dataset, termed FineRegion-CN, following the region-text annotation pipeline of FineHARD. It contains 12 million images with bounding boxes and fine-grained region descriptions in Chinese. FineRegion-CN are publicly released to support research on Chinese fine-grained multimodal understanding.}
\iclrre{An overview of our training datasets is provided in Appendix~\ref{sec:appendix_data}.}

\subsection{Bilingual Evaluation Protocols}
Existing multimodal benchmarks in English are diverse and well-established, covering a broad spectrum of vision-language tasks such as fine-grained object-level understanding (e.g., FG-OVD~\citep{bianchi2024devil}), open-vocabulary object detection (e.g., LVIS~\citep{gupta2019lvis}, COCO~\citep{lin2014microsoft}), and image-text retrieval (e.g., Flickr30K~\citep{young2014image}, DCI\cite{urbanek2024dci}). These resources enable comprehensive evaluation of model capabilities across different granularities and semantic complexities. In contrast, Chinese multimodal datasets remain limited in scope and diversity, with most focusing on short-caption retrieval tasks such as COCO-CN~\citep{li2019cococn} and Flickr30k-CNA~\citep{xie2023ccmb}. Such benchmarks are insufficient for evaluating fine-grained cross-modal alignment, particularly at the region level or with long, descriptive textual inputs.

To address this gap, we introduce a suite of Chinese evaluation benchmarks tailored for fine-grained vision-language tasks. \xie{We first construct three long-caption image-text retrieval datasets: LIT-CN, DCI-CN, and DOCCI-CN. These datasets support the evaluation of cross-modal alignment with rich and descriptive textual inputs. We then present BoxClass-CN, a region-based classification dataset designed to assess region-level vision-language alignment in Chinese.}

LIT-CN integrates diverse sources: 15,000 images from AI-Challenger Caption~\citep{aic}, 3,230 from MUGE~\citep{M6}, and 20,000 from curated web images. All images are uniformly re-captioned using Qwen2.5-VL-32B-Instruct-AWQ~\citep{bai2025qwen2.5}, prompted to generate rich, context-aware descriptions with an average length of 131 tokens. Images below 256×256 resolution are filtered, resulting in 33,010 high-quality image-text pairs. DCI-CN is derived from the Densely Captioned Images (DCI) dataset~\citep{urbanek2024dci}, with English captions translated into Chinese using the same LMM. The translations are validated by native speakers to ensure linguistic fluency and alignment with the original semantics. Similarly, DOCCI-CN is constructed from the DOCCI~\citep{onoe2024docci} dataset, following an identical translation and validation pipeline. 

BoxClass-CN is a region classification dataset that evaluates the alignment between image regions and their corresponding Chinese textual descriptions. \xie{It complements existing Chinese benchmarks by providing region-level supervision and serves as an evaluation suite for assessing models’ fine-grained understanding of visual content.} We construct this dataset through a scalable automated pipeline based on the LAION-2B corpus~\citep{schuhmann2022laion}. We first sample 200,000 images and generate detailed captions using a LMM~\citep{hong2024cogvlm2}. These captions are parsed to extract referring expressions, which are then localized using a pretrained object detector~\citep{cheng2024yolo} to produce bounding box proposals. Non-maximum suppression removes overlapping boxes, and only those with region-text similarity above 0.15 (computed by FG-CLIP~\citep{xie2025fg}) are retained. \xie{Detected categories undergo semantic clustering and merging, 
resulting in 566 semantically refined categories.} These categories are translated into Chinese, and the final dataset consists of 24,221 images and 66,258 high-quality region-text pairs. \iclrre{We provide examples of the proposed datasets in Appendix~\ref{appendix:lit-cn} and~\ref{appendix:boxclass-cn}.} Together, these datasets provide a rigorous and comprehensive assessment for bilingual vision-language alignment models, supporting deeper evaluation of fine-grained understanding capability.

\section{Experiments}
\subsection{Implementation Details}



The first stage is conducted on 160$\times$ASCEND 910B NPUs, and the second stage uses 16$\times$NVIDIA H800 GPUs. We use three vision encoder configurations: ViT-B/16, ViT-L/16, and ViT-So/16, initialized with SigLIP 2~\citep{siglip2} pre-trained weights. We employ the AdamW optimizer with a learning rate of $ 1 \times 10^{-6} $ and a weight decay coefficient of 0.001. The momentum parameters $ \beta_1 $ and $ \beta_2 $ are set to 0.9 and 0.98, respectively. A learning rate warmup strategy is applied during the first 300 iterations for stability. To accelerate training, we employ Zero-2~\citep{deepspeed}, CUDA TF32 precision, FlashAttention~\citep{dao2023flashattention}, and BFloat16 mixed-precision training. Batch sizes are set based on model size and training stage. In the first stage, the global batch sizes are 61,440 for ViT-B, 30,720 for ViT-L, and 18,432 for ViT-So. In the second stage, they are reduced to 4,096, 3,072, and 2,560, respectively. All models are trained for one epoch per stage.

\begin{table}[!t]

\begin{center}
\begin{small}
\caption{\xie{Performance comparison on fine-grained understanding tasks using Top-1 accuracy.}}
\vskip 0.1in
\label{table_fgovd}
\scalebox{0.93}{
\begin{tabular}{lccccc}
\toprule
Method & Backbone & \multicolumn{4}{c}{Fine-Grained Understanding}\\
&&Hard&Medium&Easy&Trivial\\
\midrule
CLIP&ViT-B/16&12.0&23.1&22.2&58.5\\
EVA-CLIP&ViT-B/16&14.0&30.1&29.4&58.3\\
Long-CLIP&ViT-B/16&9.2&18.4&16.2&51.8\\
FineCLIP&ViT-B/16&26.8&49.8&50.4&71.9\\
SigLIP 2&ViT-B/16&24.9&46.5&48.7&85.0\\
FG-CLIP&ViT-B/16&46.1&66.6&68.7&83.4\\
FG-CLIP 2&ViT-B/16&\textbf{52.3}&\textbf{76.3}&\textbf{80.3}&\textbf{92.0}\\
\midrule
CLIP&ViT-L/14&15.4&25.3&25.7&38.8\\
EVA-CLIP&ViT-L/14&18.3&38.4&35.2&62.7\\
Long-CLIP&ViT-L/14&9.6&19.7&16.0&39.8\\
FineCLIP&ViT-L/14&22.8&46.0&46.0&73.6\\
SigLIP 2&ViT-L/16&24.1&47.1&47.4&84.1\\
FG-CLIP&ViT-L/14&48.4&69.5&71.2&89.7\\

FG-CLIP 2&ViT-L/16&\textbf{55.5}&\textbf{77.5}&\textbf{83.1}&\textbf{92.5}\\
\midrule
Meta CLIP 2 &ViT-H/14 & 16.5&36.6&34.7&79.6 \\
SigLIP 2&ViT-So/16& 26.0&48.7&49.9&87.4\\
FG-CLIP 2&ViT-So/16  &   \textbf{54.0}&\textbf{77.4}& \textbf{79.8} &\textbf{93.5}\\
\bottomrule
\end{tabular}
}
\end{small}
\end{center}
\end{table}

\begin{table}[!t]
\caption{\xie{Performance comparison on bounding box classification tasks using Top-1 accuracy.}}
\label{table_bbox}
\begin{center}
\begin{small}
\scalebox{0.85}{
\begin{tabular}{lcccc}
\toprule
Method & Backbone & COCO\textsuperscript{80} & LVIS\textsuperscript{1203} & BoxClass-CN\textsuperscript{566} \\
\midrule
CLIP&ViT-B/16&44.2&20.9&-\\
EVA-CLIP&ViT-B/16&30.6&14.4&-\\
Long-CLIP&ViT-B/16&36.7&18.2&-\\
FineCLIP&ViT-B/16&48.4&23.3&-\\
SigLIP 2&ViT-B/16&53.4&20.6&57.9\\
FG-CLIP&ViT-B/16&52.3&28.6&-\\
FG-CLIP 2&ViT-B/16&\textbf{74.9}&\textbf{47.3}&\textbf{60.7}\\
\midrule
CLIP&ViT-L/14&33.8&9.3&-\\
EVA-CLIP&ViT-L/14&32.1&18.3&-\\
Long-CLIP&ViT-L/14&35.6&10.4&-\\
FineCLIP&ViT-L/14&54.5&22.5&-\\
SigLIP 2&ViT-L/16&54.7&25.9&56.6\\
\iclrre{CLOC}&ViT-L/14&72.9&32.6&-\\
FG-CLIP&ViT-L/14&63.2&38.3&-\\
FG-CLIP 2&ViT-L/16&\textbf{74.0}&\textbf{41.9}&\textbf{68.6}\\
\midrule
Meta CLIP 2 &ViT-H/14 & 52.0&24.4&55.2 \\
SigLIP 2&ViT-So/16& 62.0&31.4&63.6\\
FG-CLIP 2&ViT-So/16  &   \textbf{77.4}&\textbf{43.9}  &\textbf{66.5}\\
\bottomrule
\end{tabular}
}
\end{small}
\end{center}
\vskip -0.2in
\end{table}

\subsection{Localization tasks}
\subsubsection{Fine-Grained Understanding}

We evaluate open-source image-text alignment models on FG-OVD~\citep{bianchi2024devil}, a fine-grained benchmark emphasizing grounding in specific local image regions. Each region is paired with one positive description and ten synthetically perturbed negatives, forming a challenging discrimination task. The benchmark comprises four subsets: trivial, easy, medium, and hard, arranged by increasing linguistic subtlety between correct and distractor texts, requiring finer-grained reasoning for accurate matching.
Following FineCLIP~\citep{jingfineclip}, we extract dense visual features and use ROIAlign with provided bounding boxes to obtain region-specific representations. Similarity scores are computed between region features and textual descriptions, with Top-1 accuracy used as the evaluation metric. \iclrre{Results in Table~\ref{table_fgovd} show that FG-CLIP 2 achieves significant gains over prior models.} This demonstrates its superior capability in distinguishing subtle visual-linguistic correspondences, a key requirement for fine-grained understanding.

\begin{table*}[t]
\begin{center}
\begin{small}
    \caption{\xie{Open-vocabulary object detection results on LVIS\textsuperscript{minival} and LVIS. AP is reported across all categories and frequency splits.}}
    \label{table_lvis}
  \scalebox{0.88}{
  \begin{tabular}{lccccccccc}
  \toprule  
         \multirow{2}{*}{Method} &\multirow{2}{*}{Backbone}&     \multicolumn{4}{c}{LVIS\textsuperscript{minival}}    &    \multicolumn{4}{c}{LVIS} \\
         &&AP&AP\textsubscript{r}&AP\textsubscript{c}&AP\textsubscript{f}&AP&AP\textsubscript{r}&AP\textsubscript{c}&AP\textsubscript{f}\\
        \midrule
        YOLO-World-L~\citep{cheng2024yolo}&YOLOv8-L& 35.5&24.4&34.0&38.8& 28.7&22.9&24.9&35.4\\
        OWL-ST~\citep{ye2023mplugowl}&ViT-B/16&34.4& 38.3& –& –& 28.6& 30.3& –& –\\
         DetCLIP v3~\citep{yao2024detclipv3}&Swin-T & 47.0& 45.1& 47.7& 46.7& 38.9& 37.2& 37.5& 41.2\\
         T-Rex2~\citep{trex2}&Swin-T& 42.8& 37.4& 39.7& 46.5& 34.8& 29.0& 31.5& 41.2\\
         OV-DINO~\citep{ovdino}& Swin-T& 40.1& 34.5& 39.5& 41.5& 32.9& 29.1& 30.4& 37.4\\
        LLMDet&Swin-T&44.7 &37.3& 39.5& 50.7&34.9& 26.0& 30.1& 44.3\\
      \midrule
        GLIP~\citep{li2021glip}&Swin-L& 37.3 &28.2& 34.3& 41.5& 26.9& 17.1& 23.3& 36.4\\
        Grounding-DINO~\citep{liu2023groundingdino}&Swin-L& 33.9& 22.2& 30.7& 38.8& –& –& –& –\\
        OWL-ST~\citep{ye2023mplugowl}&ViT-L/14& 40.9& 41.5& –& –& 35.2& 36.2& –& –\\
        MM-GDINO~\citep{zhao2024open} & Swin-L& 36.8& 28.1& 31.8& 42.8 &29.1& 19.7& 25.6 &37.2\\
        LLMDet& Swin-L& 51.1& 45.1& 46.1& 56.6 &42.0& 31.6& 38.8 &50.2\\
      \midrule
         
          LLMDet + FG-CLIP&Swin-T + ViT-B/16&48.0&40.6&47.7&51.4   &  41.0&35.1&38.9&45.8   \\
           LLMDet + SigLIP 2&Swin-T + ViT-B/16&47.9& 42.4& 45.6& 51.0   &    41.8&36.1&39.9&45.2  \\
           LLMDet + FG-CLIP 2&Swin-T + ViT-B/16& \textbf{51.6}&\textbf{47.5}&\textbf{50.7}&\textbf{53.2}  &\textbf{44.0}&\textbf{37.4}&\textbf{42.8}&\textbf{48.2}   \\
           
        \midrule
          LLMDet + FG-CLIP&Swin-T + ViT-L/14& 50.5& 41.9&49.3&53.1 &43.1&37.0&41.3&47.7\\
           LLMDet + SigLIP 2&Swin-T + ViT-L/16& 49.9&46.9&48.9&51.3&43.6&40.4&42.0&45.9\\
      LLMDet + FG-CLIP 2&Swin-T + ViT-L/16&  \textbf{52.6}&\textbf{48.6}&\textbf{51.8}&\textbf{54.0}&\textbf{45.5}&\textbf{41.0}&\textbf{44.2}&\textbf{49.0}\\
      \midrule
        LLMDet + SigLIP 2&Swin-T + ViT-So/16&  50.2&48.4&49.0&51.7 &44.3&41.1&42.7&46.4 \\
      LLMDet + Meta CLIP 2&Swin-T + ViT-H/14&  52.2&50.5&51.1&53.5 & 44.5&41.7&42.9&47.6\\
      LLMDet + FG-CLIP 2&Swin-T + ViT-So/16&  \textbf{53.1}&\textbf{50.8}&\textbf{52.3}&\textbf{54.2}&\textbf{45.9}&\textbf{42.1}&\textbf{44.6}&\textbf{49.0}\\
     \bottomrule
  \end{tabular}
  }

\end{small}
\end{center}
\end{table*}

\begin{table*}[!tbp]
\caption{Comparisons on English image-level tasks, including long/short caption image-text retrieval, and zero-shot image classification. }
\label{table_retrieval_en}
\begin{center}
\begin{small}
\scalebox{0.88}{
\begin{tabular}{lccccccccccc}
\toprule
\multirow{2}{*}{Method} & \multirow{2}{*}{Backbone} 
& \multicolumn{2}{c}{ShareGPT4V} & \multicolumn{2}{c}{DCI} 
& \multicolumn{2}{c}{MSCOCO}& \multicolumn{2}{c}{Flickr30k}
&IN-1K&IN-v2\\
&&I$\rightarrow$T&T$\rightarrow$I&I$\rightarrow$T&T$\rightarrow$I&I$\rightarrow$T&T$\rightarrow$I&I$\rightarrow$T&T$\rightarrow$I&Top-1&Top-1 \\
\midrule
CLIP&ViT-B/16&78.2&79.6&45.5&43.0&51.8&32.7&82.2& 62.1&68.4&61.9\\
EVA-CLIP&ViT-B/16&90.5&85.5&41.9&41.2&58.7&41.6&85.7&71.2&74.7&67.0\\
Long-CLIP&ViT-B/16&94.7&93.4&51.7&57.3&57.6&40.4&85.9&70.7 &66.8&61.2\\
FineCLIP&ViT-B/16&70.6&73.3&35.5&34.4&54.5&40.2&82.5&67.9&55.7&48.8\\
\iclrre{UMG-CLIP}&ViT-B/16&-&-&-&-&64.7&51.6&91.4&78.6&-&66.5\\
FG-CLIP&ViT-B/16&\textbf{96.7}&94.9&61.8&60.6&64.1&45.4&90.7&76.4&69.0&61.8\\
SigLIP 2&ViT-B/16&   66.0&67.9&32.3&34.2&71.2&\textbf{55.2}&92.6&78.0&\textbf{81.2} &\textbf{74.5}   \\
FG-CLIP 2&ViT-B/16&95.8&\textbf{95.4}&\textbf{64.5}&\textbf{64.9}&\textbf{72.1}&54.5&\textbf{94.1}&\textbf{81.9}&79.5&72.2\\
\midrule
CLIP&ViT-L/14&86.5&83.6&37.2&36.4&58.0&37.1&87.4&67.3&76.6&70.9\\
EVA-CLIP&ViT-L/14&91.5&89.4&47.2&47.8&64.2&47.9&89.2&77.9&80.4&73.8\\
Long-CLIP&ViT-L/14&95.8&95.6&44.2&52.5&62.8&46.3&90.0&76.2&73.5&67.9\\
FineCLIP&ViT-L/14&73.4&82.7&40.1&46.2&-&-&-&-&60.8&53.4\\
\iclrre{UMG-CLIP}&ViT-L/14&-&-&-&-&68.9&54.6&93.4&83.1&-&71.6\\
FG-CLIP&ViT-L/14&\textbf{97.4}&\textbf{96.8}&66.7&66.1&68.9&50.9&93.7&81.5&76.1&69.0\\
\iclrre{CLOC}&ViT-L/14& -&-&-&-&74.8&54.4&-&-&80.1&73.2     \\
SigLIP 2&ViT-L/16& \green{85.1}&\green{84.6}&\green{48.0}&\green{49.3}&\green{72.1}&\green{55.2}&\green{94.3}&\green{82.6}&\textbf{\green{83.1}}& \green{76.5}    \\
FG-CLIP 2&ViT-L/16  &96.9&96.6&\textbf{70.0}&\textbf{71.6}&\textbf{75.1}&\textbf{58.6}&\textbf{96.6}&\textbf{84.8}&\green{83.0}&\textbf{77.4}  \\

\midrule
SigLIP 2&ViT-So/16& 78.6&79.5&46.0&47.1&\green{71.0}&\green{55.8}&94.1&82.5&\green{83.8} &\green{77.7}   \\

Meta CLIP 2 &ViT-H/14  &93.9&89.2&53.0&50.2&66.8&47.7&91.9&77.0&81.7&75.7   \\

FG-CLIP 2&ViT-So/16  &\textbf{97.5}&\textbf{96.7}&\textbf{70.6}&\textbf{72.1}&\textbf{74.6}&\textbf{56.7}&\textbf{95.9}&\textbf{85.0}&\textbf{84.1}&\textbf{77.8}     \\
\bottomrule
\end{tabular}
}
\end{small}
\end{center}
\vskip -0.2in
\end{table*}

\begin{table*}[!ht]
\caption{Performance on Chinese image-text retrieval benchmarks, covering both long-text and short-text settings. Results are reported in terms of Recall@1 (\%).}
\label{table_retrieval_cn}
\begin{center}
\begin{small}
\scalebox{0.9}{
\begin{tabular}{lccccccccccc}
\toprule
\multirow{2}{*}{Method} & \multirow{2}{*}{Backbone} 
& \multicolumn{2}{c}{LIT-CN} & \multicolumn{2}{c}{DCI-CN} 
& \multicolumn{2}{c}{DOCCI-CN}& \multicolumn{2}{c}{Flickr-CNA}& \multicolumn{2}{c}{COCO-CN}\\
&&I$\rightarrow$T&T$\rightarrow$I&I$\rightarrow$T&T$\rightarrow$I&I$\rightarrow$T&T$\rightarrow$I&I$\rightarrow$T&T$\rightarrow$I&I$\rightarrow$T&T$\rightarrow$I \\
\midrule
R2D2&ViT-B/16& 35.7&27.4&25.9&27.3&36.1&36.9&69.7&51.1&60.1&45.5  \\
Chinese-CLIP&ViT-B/16& 45.7&35.6&30.1&27.9&44.6&43.1&75.8&62.4&68.8&54.9 \\
SigLIP 2&ViT-B/16&  \green{4.6}&\green{2.6}&5.0&\green{4.0}&7.6&5.7&71.7&49.1&\green{68.6}&46.2 \\

FG-CLIP 2&ViT-B/16&\textbf{82.4}&\textbf{81.1}&\textbf{53.9}&\textbf{55.7}&\textbf{71.2}&\textbf{75.4}&\textbf{85.4}&\textbf{69.9}&\textbf{77.2}&\textbf{62.9}\\
\midrule
R2D2&ViT-L/14& 48.3&33.3&35.6&34.2&49.5&46.3&78.8&60.0&69.6&52.7 \\
Chinese-CLIP&ViT-L/14& 48.6&38.9&31.4&32.7&49.7&50.8&82.9&69.6&74.3&59.9\\
SigLIP 2&ViT-L/16 & \green{14.8}&\green{10.9}&\green{13.6}&\green{14.4}&\green{24.6}&\green{27.3}&\green{79.8}&\green{53.2}&\green{74.2}&\green{51.7}  \\

FG-CLIP 2&ViT-L/16& \textbf{86.3}&\textbf{85.9}&\textbf{60.4}&\textbf{62.2}&\textbf{77.6}&\textbf{81.9}&\textbf{90.3}&\textbf{75.0}&\textbf{82.8}&\textbf{66.5}   \\
\midrule

SigLIP 2&ViT-So/16&\green{17.0}&\green{10.8}&13.4&12.0&25.0&21.3&78.4&51.7&72.0&50.7     \\
Meta CLIP 2 &ViT-H/14 &77.2&67.6&53.8&52.1&73.8&77.2&89.3&72.2&80.1&63.1    \\
FG-CLIP 2&ViT-So/16   &\textbf{87.6}&\textbf{86.3}&\textbf{62.7}&\textbf{65.1}&\textbf{79.7}&\textbf{84.0}&\textbf{91.5}&\textbf{77.2}&\textbf{83.2}&\textbf{68.1}      \\
\bottomrule
\end{tabular}
}
\end{small}
\end{center}
\end{table*}

\subsubsection{Bounding Box Classification}
We evaluate zero-shot bounding box classification on COCO-val2017~\citep{lin2014microsoft}, LVIS~\citep{gupta2019lvis}, and our proposed BoxClass-CN dataset, following the protocol of~\citep{xie2025fg}. While COCO and LVIS focus on English category recognition within localized regions, BoxClass-CN targets Chinese, enabling a bilingual assessment of fine-grained vision-language alignment. \iclrre{As shown in Table~\ref{table_bbox}}, FG-CLIP 2 achieves state-of-the-art performance in both languages and significantly outperforms all compared open-source models. These results demonstrate its robust ability to align local visual content with semantic concepts across linguistic boundaries.

\subsubsection{Open-Vocabulary Object Detection}
To assess the impact of vision-language alignment models on open-vocabulary detection (OVD), we adopt a training-free evaluation protocol that avoids biases from detector fine-tuning. Unlike prior approaches~\citep{wu2023clipself} that require small-scale detector training on fixed categories, our method directly leverages the zero-shot generalization of alignment models to calibrate confidence scores and category predictions in the final OVD output. This allows for a cleaner analysis of the contribution of image-text alignment to OVD performance. We adopt a fusion strategy that leverages the vision-language alignment model to refine the class predictions and confidence scores of a pre-trained detector. Specifically, similarity scores from the alignment model are combined with the detector’s original confidences via geometric averaging, enabling more semantically accurate and calibrated open-vocabulary detection.
We use LLMDet~\citep{fu2025llmdet} as the base detector and evaluate on the challenging LVIS dataset~\citep{gupta2019lvis}, which contains 1,203 categories. As shown in Table~\ref{table_lvis}, FG-CLIP 2 combined with LLMDet achieves the best performance among open-source methods, demonstrating its strong practical utility and  superior generalization in detection scenarios.

\subsection{Image-level tasks}

\subsubsection{Long/Short Caption Image-Text Retrieval}
To comprehensively evaluate image-text alignment under varying linguistic complexity, we conduct experiments on both short and long caption retrieval tasks. Short-text retrieval (Flickr30k~\citep{young2014image}, MSCOCO~\citep{lin2014microsoft}) assesses basic semantic matching, while long-text retrieval requires fine-grained understanding of detailed descriptions and complex visual scenes.
For short-text retrieval, we employ the validation set of MSCOCO and the test set of Flickr30k, which are widely used benchmarks for assessing image-text alignment models.
For long-text retrieval in English, we use the 1K subset of ShareGPT4V~\citep{chen2025sharegpt4v} and the full test set of DCI~\citep{urbanek2024dci} following the protocol of Long-CLIP~\citep{zhang2025long}. For Chinese, we evaluate on LIT-CN, DCI-CN, and DOCCI-CN for long-text, and Flickr-CNA~\citep{xie2023ccmb} and COCO-CN~\citep{li2019cococn} for short-text. 
We employ the validation set of COCO-CN and the test set of Flickr-CNA.
These datasets cover diverse content and caption styles, providing a robust evaluation of multilingual performance. As shown in Table~\ref{table_retrieval_en} and Table~\ref{table_retrieval_cn}, FG-CLIP 2 achieves consistent improvements across all settings, with particularly strong gains on long-text retrieval, highlighting its superior capability in fine-grained vision-language alignment. 
\TODO{Notably, FG-CLIP 2 outperforms Meta CLIP 2~\citep{chuang2025metaclip2}, the current multilingual SOTA, on both language settings, despite using a smaller ViT-L/16 backbone with 1.0 billion parameters compared to Meta CLIP 2's 1.8 billion parameter ViT-H/14. This highlights the effectiveness of our training paradigm in achieving stronger performance with reduced model scale.}

\begin{table}[!tbp]
\begin{center}
\begin{small}
\caption{\xie{Performance on dense prediction tasks.}}
\label{table_ovseg}
\scalebox{0.7}{
\begin{tabular}{lccccccc}
\toprule
Model& Backbone& A-847& PC-459& A-150& PC-59& VOC-20& VOC-21\\
\midrule
CLIP& ViT-B/16& 8.4& 16.6& 27.2& 57.5& 93.7& 78.3\\
CLIPSelf& ViT-B/16& 10.1& -& 29.7& 55.3& -& -\\
FineCLIP& ViT-B/16& 12.2& -& 32.4& 56.0& -& -\\
\iclrre{UMG-CLIP}& ViT-B/16& 13.8&21.1 & 34.6& 58.2&- &- \\
FG-CLIP& ViT-B/16& 12.3 &19.1&33.4&58.2&95.3&77.9\\ 
SigLIP 2& ViT-B/16&10.4&17.0 &28.5&55.4&94.4&75.8 \\
FG-CLIP 2& ViT-B/16&\textbf{16.6}&\textbf{24.0}&\textbf{38.5} & \textbf{61.2}&\textbf{97.1}&\textbf{81.1} \\

\midrule
CLIP& ViT-L/14& 10.8& 20.4& 31.5& 62.0& 96.6& 81.8\\
CLIPSelf& ViT-L/14& 13.6& -& 34.9& 59.1& -& -\\
FineCLIP& ViT-L/14& 14.1& -& 36.1& 59.9& -& -\\
\iclrre{UMG-CLIP}& ViT-L/14& 15.4&23.2 & 36.1& 58.7&- &- \\
FG-CLIP& ViT-L/14& 14.6&23.3 & 36.9& 61.4&97.4 &81.8 \\
SigLIP 2& ViT-L/16& 14.3& 24.1& 38.8 &62.4 &97.0& 82.3\\
FG-CLIP 2& ViT-L/16&18.8&26.6&41.2&62.4&97.6&81.8 \\
FG-CLIP 2& ViT-So/16&\textbf{20.0}&\textbf{27.5}&\textbf{42.2}&\textbf{63.3}&\textbf{97.8}&\textbf{83.2} \\

\bottomrule
\end{tabular}
}
\end{small}
\end{center}
\end{table}


\begin{table*}[htbp]
\caption{Comparisons on large multimodal model benchmarks.}
\begin{center}
\scalebox{0.8}{
\begin{tabular}{lccccccccccccc}
\toprule
\multirow{2}{*}{Method} & \multirow{2}{*}{GQA} & \multirow{2}{*}{MMMU} &\multirow{2}{*}{TextVQA} &\multicolumn{3}{c}{RefCOCO} &\multicolumn{2}{c}{MMBench-EN} &\multicolumn{2}{c}{MMBench-CN} &\multicolumn{3}{c}{V* Benchmark}\\
&&&&Val&TestA&TestB&Dev&Test&Dev&Test&AR&SRR&Overall \\
\toprule  
LLaVA-1.5 + CLIP&61.9&35.7&58.2&76.2&83.4&67.9& 65.1 & 66.5 & 58.2 & 58.4 &44.4 &52.6 &47.6\\
LLaVA-1.5 + SigLIP 2&62.0&37.0&56.6&79.5&84.4&73.9& 66.2 & 64.8 & 58.0 & 57.2 &40.9&56.6&47.1\\
LLaVA-1.5 + Meta CLIP 2&62.4&37.0&59.4&76.7&82.8&69.8& 66.5 & \textbf{67.0} & 59.3 & 60.3&39.1&54.0&45.0\\
LLaVA-1.5 + FG-CLIP 2&\textbf{64.0}&\textbf{38.1}&\textbf{62.0}&\textbf{84.9}&\textbf{89.8}&\textbf{79.5}& \textbf{67.6} & 66.5 & \textbf{60.5} & \textbf{61.4}& \textbf{45.2}& \textbf{57.9}& \textbf{50.3}\\

\bottomrule
\end{tabular}
}
\end{center}
\label{tab:table_lmm}
\end{table*}

\begin{table*}[!tbp]
  \centering
  \small              
  \caption{Ablation study results for the training objectives of FG-CLIP 2.}
  \scalebox{0.95}{
  \begin{tabular}{lcccccccccc}
  \toprule           
         \multirow{2}{*}{Method} & 
         \multicolumn{2}{c}{COCO\textsuperscript{80}} & 
         \multicolumn{4}{c}{FG-OVD} & 
         \multicolumn{2}{c}{Flickr30k}&
         \multicolumn{1}{c}{ImageNet-V2} & \\
&\multicolumn{1}{c}{Top-1}&Top-5&Hard&Medium&Easy&Trivial&I$\rightarrow$T&T$\rightarrow$I&Top-1\\
\midrule
    FG-CLIP 2   &74.9&95.7&52.3&76.3&80.3&92.0&94.1& 81.9& 72.2\\
    \midrule
    ~~w/o $\mathcal{L}_{\text{CMR}}$   &74.0&94.9&50.9&75.1&77.8&93.5&93.3&81.9&72.1\\
    ~~w/o $\mathcal{L}_{\text{TIC}}$  &70.1&94.8&51.6&75.1&79.1&92.1&93.7&81.8&72.1\\
    ~~w/o $\mathcal{L}_{\text{CMR}}$, $\mathcal{L}_{\text{TIC}}$   &62.7&93.4&51.7&76.0&77.6&90.7&93.5&81.6&72.0\\
     \bottomrule
  \end{tabular}
  }
   \label{table:albation}
\end{table*}

\subsubsection{Zero-shot Image Classification}
We evaluate zero-shot image classification on ImageNet-1K~\citep{deng2009imagenet} and ImageNet-v2~\citep{recht2019imagenet} using standard prompts~\citep{openaiclip}. \TODO{As shown in Table~\ref{table_retrieval_en}, FG-CLIP 2 achieves competitive performance compared to SigLIP 2~\citep{siglip2}, and outperforms EVA-CLIP~\citep{sun2023eva}, Long-CLIP~\citep{zhang2025long}, FineCLIP~\citep{jingfineclip}, \iclrre{UMG-CLIP~\citep{shi2024umg}, CLOC~\citep{chen2024contrastive}, and Meta CLIP 2.}} This confirms that the improvements in fine-grained alignment do not come at the cost of standard recognition accuracy, demonstrating a well-balanced representation capability.


\subsection{Dense prediction tasks}
We evaluate dense prediction through open-vocabulary segmentation, a task that requires models to segment object categories beyond a fixed training set. We adopt Cat-Seg~\citep{cho2024catseg} as the base framework, trained on COCO-Stuff-164k (172 categories), and evaluate on datasets with diverse category schemas: ADE20k (847 or 150 categories, denoted A-847/A-150), Pascal Context (PC-459/PC-59), and Pascal VOC (VOC-20/VOC-21). 
As shown in Table~\ref{table_ovseg}, FG-CLIP~2 achieves the best performance across models of various scales, delivering consistent improvements over the baseline. This demonstrates its strong capability in enabling pixel-level generalization, crucial for downstream dense understanding tasks.

\subsection{Large Multimodal Model tasks}
We empoly FG-CLIP 2 as a vision encoder in large multimodal models (LMMs) to assess its compatibility and utility in advanced multimodal reasoning. We integrate FG-CLIP~2 into a standard LLaVA-style LMM architecture, following the pre-training and supervised fine-tuning protocol of LLaVA-1.5~\citep{llava}. 
\final{Evaluation is conducted on a diverse set of benchmarks spanning visual question answering (GQA~\citep{hudson2019gqa}, TextVQA~\citep{textvqa}), multimodal understanding (MMMU~\citep{yue2023mmmu}), general multimodal evaluation (MMBench-EN and MMBench-CN~\citep{liu2024mmbench}), referring expression comprehension (RefCOCO~\citep{refcoco}), and fine-grained visual reasoning (V* Benchmark~\citep{wu2024v}).}
Results in Table~\ref{tab:table_lmm} show that LMMs equipped with FG-CLIP 2 outperform those using other open-source encoders, including SigLIP 2 and Meta CLIP 2. This indicates that the fine-grained and bilingual capabilities of FG-CLIP 2 effectively transfer to higher-level multimodal tasks, making it a strong candidate for integration into next-generation LMMs.

\subsection{Ablation Study}
\xie{As our training framework leverages global-level, region-level, and hard-negative data through complementary learning mechanisms, we adopt a baseline model that incorporates global alignment alongside fine-grained visual and textual learning. This allows us to specifically evaluate the impact of the Cross-modal Rank Loss with Global Threshold Synchronization ($\mathcal{L}_{\text{CMR}}$) and the Textual Intra-modal Contrastive Loss ($\mathcal{L}_{\text{TIC}}$).}
As shown in Table~\ref{table:albation}, removing $\mathcal{L}_{\text{TIC}}$ leads to a 4.8-point drop in COCO Top-1 accuracy (to 70.1\%) and a decrease in FG-OVD Hard performance from 52.3\% to 51.6\%, confirming its critical role in distinguishing semantically similar texts. Removing $\mathcal{L}_{\text{CMR}}$ also degrades performance, with FG-OVD Hard falling to 50.9\%, indicating its importance in cross-modal alignment. When both losses are removed, COCO Top-1 drops sharply to 62.7\%, demonstrating their complementary benefits. The full model achieves consistent gains across all tasks, \iclrre{especially in bounding box classification and fine-grained understanding, while maintaining strong results on standard recognition benchmarks,} validating the effectiveness of our proposed training objectives. 
\iclrre{Additionally, visualizations demonstrating the semantic separability improvement from TIC loss are provided in Appendix~\ref{appendix:tic}.}

\iclrre{To examine how different data compositions affect the model’s bilingual capability, we conduct an ablation study comparing two variants trained under identical settings except for the data used in the second stage: one uses English-only data, while the other uses both English and Chinese data. Results in Table~\ref{tab:ablation_data_composition} (Appendix~\ref{appendix:ablation_data}) show that the bilingual variant not only achieves higher accuracy on Chinese benchmarks but also consistently improves performance on English-only evaluation sets, confirming a mutually promoting effect between the two languages. We further conduct a comparison with re-trained baselines under identical training data in Appendix~\ref{app:icml_controlled_baselines}, showing that FG-CLIP 2’s architecture and learning objectives contribute significantly to its performance gains.}

\section{Conclusion and Limitations}
\label{conclusion}
In this work, we present FG-CLIP 2, a bilingual vision-language model that advances fine-grained understanding for both English and Chinese. Our two-stage training paradigm progressively refines alignment by leveraging both short and long captions, region-text supervision, and multiple discriminative objectives, including the proposed Textual Intra-modal Contrastive (TIC) loss to better distinguish semantically similar descriptions. Trained on large-scale, high-quality English and Chinese datasets, FG-CLIP~2 achieves \xie{superior} performance \TODO{across 29 datasets and 8 tasks}, demonstrating strong bilingual generalization. To advance evaluation in non-English settings, we introduce a new benchmark for Chinese multimodal understanding with challenging tasks such as long caption image-text retrieval and bounding box classification. We release the model, code, and benchmark to support future research in bilingual fine-grained vision-language understanding. In future work, we will extend the model to handle longer textual inputs and explicitly model relational structures among objects.



\section*{Impact Statement}

This paper aims to advance the field of Machine Learning, which has broad implications for society. There are many potential societal consequences of our work, none which we feel must be specifically highlighted here. 

\nocite{langley00}

\bibliography{example_paper}
\bibliographystyle{icml2026}

\newpage
\appendix
\onecolumn

\section{Training Data Details}
\label{sec:appendix_data}

\begin{figure*}[!htb]
    \renewcommand\thefigure{A}
  \centering   \includegraphics[width=1.0\linewidth]{{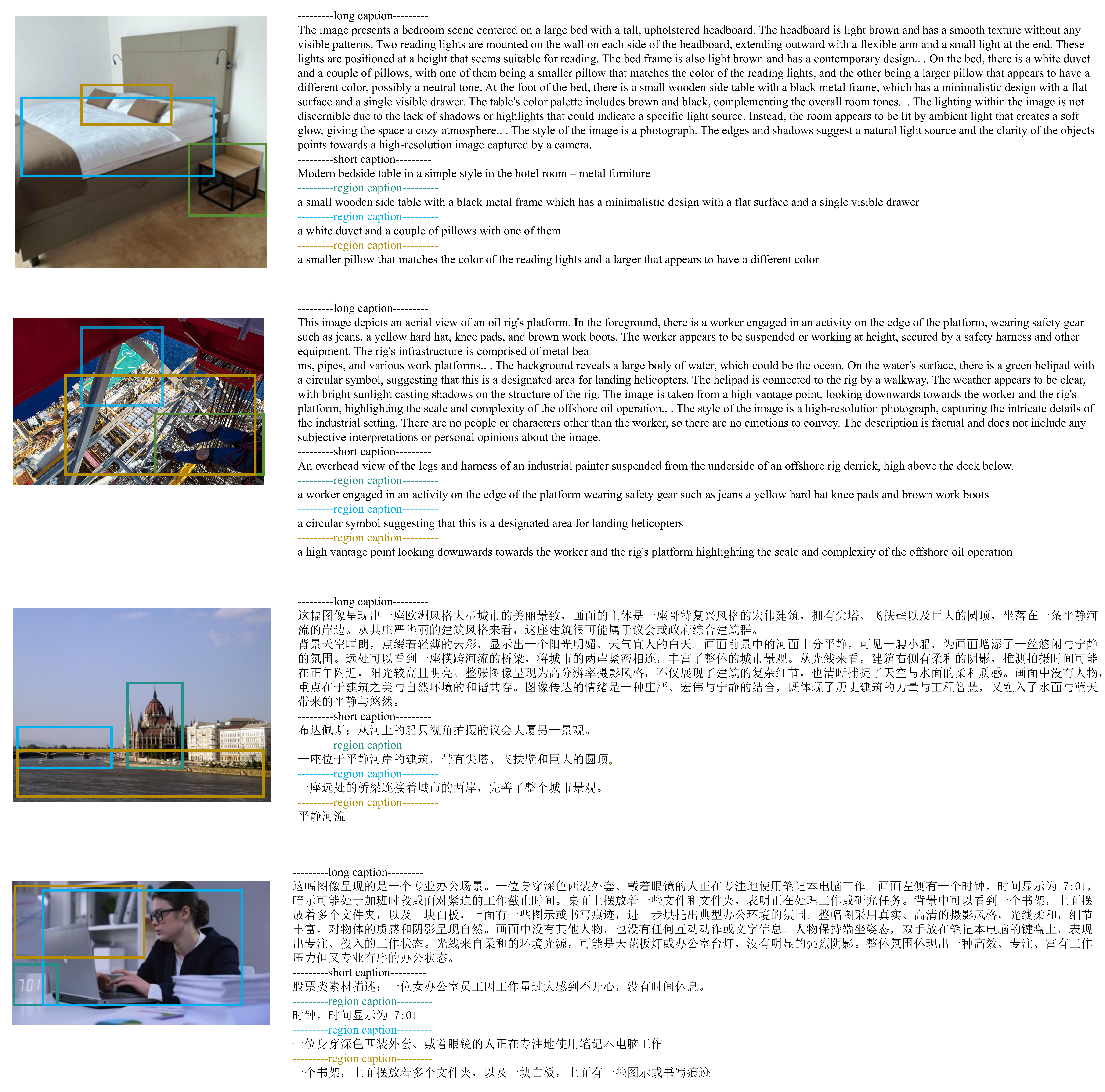}}
    \caption{\iclrre{Examples of Bilingual Training Data.}}
    
    \label{fig:cnsample}
\end{figure*}

Table~\ref{tab:training_data} summarizes the datasets used in our two-stage training pipeline. \iclrre{We present examples of the Chinese and English training data used in Stage II in Figure~\ref{fig:cnsample}.}

\begin{table}[h]
\renewcommand\thetable{A}
  \centering
  \caption{Overview of the training datasets used in two stages.}
  \resizebox{0.5\textwidth}{!}{
    \begin{tabular}{l l c r}
    \toprule
    \textbf{Data Type} & \textbf{Dataset} & \textbf{Language} & \textbf{Size} \\
    \midrule
    \multirow{4}{*}{Image-text} 
        & LAION-2B-enhanced & English & 1.6B \\
        & Wukong            & Chinese & 100M \\
        & Zero              & Chinese & 250M \\
        & \lightgray{In-house Data}  & \lightgray{Chinese} & \lightgray{500M} \\
    \midrule
    \multirow{2}{*}{Region-text} 
        & FineHARD              & English & 12M \\
        & \lightgray{FineRegion-CN} & \lightgray{Chinese} & \lightgray{12M}\\
    \bottomrule
    \end{tabular}
  }
  \label{tab:training_data}
\end{table}

\begin{table*}[!htb]
\renewcommand\thetable{B}
  \centering
  \caption{Detailed parameters for training with Cat-Seg.}
  \resizebox{0.8\textwidth}{!}{
    \begin{tabular}{c c c c}
    \toprule
    \textbf{Parameter name} & \textbf{Value} & \textbf{Parameter name} & \textbf{Value} \\
    \midrule
Text Guidance Proj Dim  & 128  & Min Size Train & 384 \\
Appearance Guidance Proj Dim  & 128       & Min Size Train Sampling      & Choice \\
Decoder Dims                  & [64, 32]  & Min Size Test                & 640 \\
Decoder Guidance Dims         & [256, 128]& Size Divisibility            & 384 \\
Decoder Guidance Proj Dims    & [32, 16]  & Format                       & RGB \\
Num Layers                    & 2         & Dataset Mapper Name          & Mask Former Semantic \\
Num Heads                     & 4         & Images Per Batch             & 8 \\
Hidden Dims                   & 128       & LR Scheduler Name            & Warmup Cosine LR \\
Pooling Sizes                 & [2, 2]    & Base Learning Rate           & 0.0002 \\
Feature Resolution            & [24, 24]  & Max Iterations               & 80,000 \\
Window Sizes                  & 12        & Backbone Multiplier          & 0.0 \\
Attention Type                & Linear    & CLIP Multiplier                & 0.01 \\
        
    \bottomrule
    \end{tabular}
  }
  \label{tab:segexpdet}
\end{table*}

    

\section{Open-Vocabulary Object Detection Experimental Details}

\xie{In our experiments, we adopt LLMDet as the base open-vocabulary object detection model. Its output consists of bounding boxes, each associated with a predicted category and a confidence score. To improve category accuracy, we recalibrate these predictions using vision-language alignment models, such as FG-CLIP 2, without modifying LLMDet’s parameters.}

\xie{For each detected bounding box, we extract its visual feature by applying RoI-Align on the dense ViT feature map produced by FG-CLIP 2. We also encode all candidate category names into text embeddings using the same model. We then compute the cosine similarity between the region’s visual feature and each category’s text embedding, followed by Softmax normalization to obtain a category-wise alignment similarity distribution. To produce the final prediction, we combine LLMDet’s original confidence score with FG-CLIP 2’s normalized similarity score via geometric averaging. The resulting fused score reflects both the detector’s localization confidence and the alignment model’s semantic relevance. We then select the category with the highest fused score as the final predicted class, and assign the corresponding fused value as the final confidence output.}

\xie{This approach leverages FG-CLIP 2’s fine-grained understanding to recalibrate LLMDet’s predictions across the entire category space. It ensures that categories with strong semantic alignment but initially low detection scores can still be correctly selected if their fused confidence is highest. This global recalibration significantly improves performance on novel categories, while maintaining compatibility with the original detector’s structure.}

\section{Dense Prediction Tasks Experimental Details}
We adopt Cat-Seg~\citep{cho2024catseg} as the base model for open-vocabulary segmentation, which supports plug-and-play integration of various image-text models. A unified training configuration is used across different ViT backbones, with detailed hyperparameters provided in Table~\ref{tab:segexpdet}.


\section{\iclrre{Ablation on Bilingual Data Composition}}
\label{appendix:ablation_data}

\iclrre{We present the results of our ablation study on bilingual data composition in Stage II. Both models are initialized from the same Stage I checkpoint trained on the full bilingual (English and Chinese) dataset. In Stage II, we compare two settings: (1) English-only training data, and (2) bilingual English-Chinese training data. As shown in Table~\ref{tab:ablation_data_composition}, the bilingual variant not only outperforms the English-only variant on Chinese benchmarks but also consistently improves performance on English-only evaluation sets, confirming a mutually promoting effect between the two languages.}


\begin{table*}[!tbp]
\renewcommand\thetable{C}
\centering
\caption{Ablation study on data composition for bilingual capability. FG-CLIP~2$^*$ (marked with $^*$) uses English-only data in Stage~II, while FG-CLIP~2 uses both English and Chinese data.}
\label{tab:ablation_data_composition}
\setlength{\tabcolsep}{5pt}
\scalebox{0.87}{
\begin{tabular}{lccccccc}
\toprule
Method & Backbone &
DCI &
MSCOCO &
DOCCI-CN &
Flickr30k-CNA &
COCO$^{80}$ &
Fine-Grained Understanding \\
& &
I$\rightarrow$T / T$\rightarrow$I &
I$\rightarrow$T / T$\rightarrow$I &
I$\rightarrow$T / T$\rightarrow$I &
I$\rightarrow$T / T$\rightarrow$I &
Top-1 / Top-5 &
Hard / Medium / Easy / Trivial \\
\midrule
FG-CLIP~2$^*$ & ViT-B/16 &
64.4 / 64.7 &
71.4 / 54.5 &
57.8 / 58.6 &
84.5 / 68.0 &
71.3 / 95.2 &
52.2 / 75.9 / 80.1 / 91.0 \\

FG-CLIP~2 & ViT-B/16 &
64.5 / 64.9 &
72.1 / 54.5 &
71.2 / 75.4 &
85.4 / 69.9 &
74.9 / 95.7 &
52.3 / 76.3 / 80.3 / 92.0 \\
\bottomrule
\end{tabular}}
\vskip -0.1in
\end{table*}

\section{\iclrre{Comparison with Re-trained Baselines}}
\label{app:icml_controlled_baselines}

\iclrre{To ensure a fair comparison and isolate the impact of model architecture and learning objectives from that of training data, we re-train two representative baselines, SigLIP~2 and Chinese-CLIP, using exactly the same Stage II training data as FG-CLIP~2. Due to computational constraints, all models in this study, including FG-CLIP~2, are trained solely on the Stage~II dataset without any Stage I pretraining. This setup guarantees identical data conditions across all methods, enabling a fully controlled evaluation focused solely on architectural and objective-level differences.}

\iclrre{Table~\ref{tab:icml_controlled_baselines} summarizes the results under this setting. Remarkably, even in the absence of large-scale pretraining, FG-CLIP~2 consistently outperforms both SigLIP~2 and Chinese-CLIP across all benchmarks, including DCI, MSCOCO, DOCCI-CN, Flickr30k-CNA, COCO$^{80}$, and Fine-Grained Understanding. These results indicate that, beyond the benefits of bilingual training data, the architectural design and tailored learning objectives of FG-CLIP~2 play a crucial role in its superior performance. The consistent gains across diverse benchmarks demonstrate that these modeling choices substantially enhance the model’s ability to capture fine-grained semantics and align cross-modal representations effectively.}

\begin{table*}[ht]
\renewcommand\thetable{D}
\centering
\caption{Comparison with retrained baselines using only Stage~II data (no Stage~I pretraining, indicated by \textdagger).}
\label{tab:icml_controlled_baselines}
\setlength{\tabcolsep}{5pt}
\scalebox{0.86}{
\begin{tabular}{lccccccc}
\toprule
Method & Backbone &
DCI &
MSCOCO &
DOCCI-CN &
Flickr30k-CNA &
COCO$^{80}$ &
Fine-Grained Understanding \\
& &
I$\rightarrow$T / T$\rightarrow$I &
I$\rightarrow$T / T$\rightarrow$I &
I$\rightarrow$T / T$\rightarrow$I &
I$\rightarrow$T / T$\rightarrow$I &
Top-1 / Top-5 &
Hard / Medium / Easy / Trivial \\
\midrule
Chinese-CLIP\textdagger & ViT-B/16 &
26.9 / 25.5 &
47.3 / 31.7 &
48.3 / 47.7 &
80.0 / 65.5 &
41.4 / 69.0 &
16.3 / 33.0 / 37.1 / 73.9 \\

SigLIP~2\textdagger & ViT-B/16 &
49.1 / 49.0 &
71.5 / 53.5 &
50.5 / 49.4 &
81.0 / 52.7 &
53.7 / 79.7 &
25.5 / 50.6 / 59.2 / 83.4 \\

FG-CLIP~2\textdagger & ViT-B/16 &
\textbf{60.9 / 62.6} &
\textbf{72.1 / 53.7} &
\textbf{65.9 / 69.5} &
\textbf{84.3 / 66.1} &
\textbf{71.0 / 95.4} &
\textbf{51.9 / 76.0 / 80.5 / 90.6} \\
\bottomrule
\end{tabular}}

 \vskip -0.1in
\end{table*}

\section{\iclrre{Hyperparameter Ablation Study}}
\label{appendix:hyper}

\iclrre{In addition to the global alignment learning objective, our model incorporates four auxiliary losses: Fine-Grained Visual Learning (FGV), Fine-Grained Textual Learning (FGT), Textual Intra-modal Contrastive Loss (TIC), and Cross-Modal Rank Loss (CMR). The weights for these losses are set using a sequential tuning strategy: starting from the base model trained with only the global alignment loss, we introduce each auxiliary loss one at a time, tune its weight while keeping previously added losses fixed (and excluding any future ones), and select the value that yields the highest average performance on the validation set.}

\begin{figure*}[!tbp]
    \renewcommand\thefigure{B}
  \centering   \includegraphics[width=1.0\linewidth]{{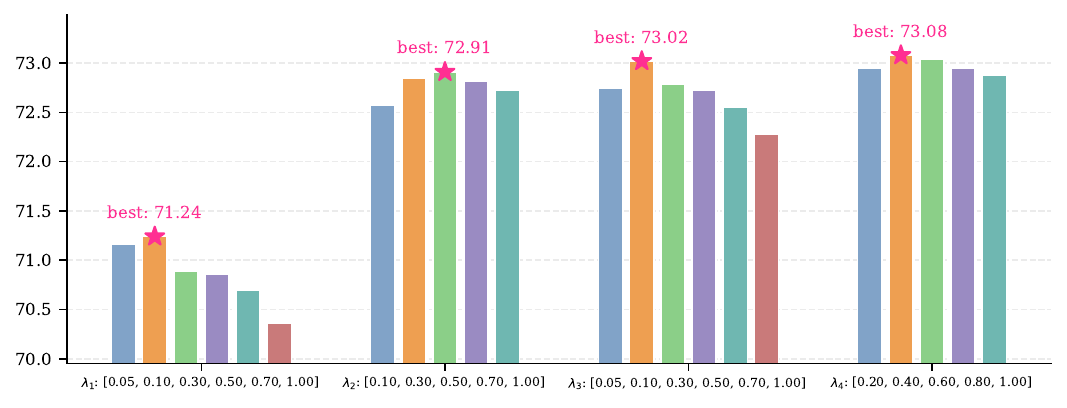}}
    \caption{\iclrre{The performance over different values of each loss hyperparameter.}}
    
    \label{fig:hyper_as}
\end{figure*}

\iclrre{Figure~\ref{fig:hyper_as} shows the average performance over different values of each loss hyperparameter, where the average is computed across several representative evaluation tasks: DCI, MSCOCO, DOCCI-CN, Flickr30k-CNA, Bbox Classification (COCO\textsuperscript{80}), and Fine-Grained Understanding.}

\section{Visualization of Semantic Alignment Capability for Dense Visual Features}

\begin{figure*}[!htbp]
    \renewcommand\thefigure{C}
  \centering   \includegraphics[width=0.8\linewidth]{{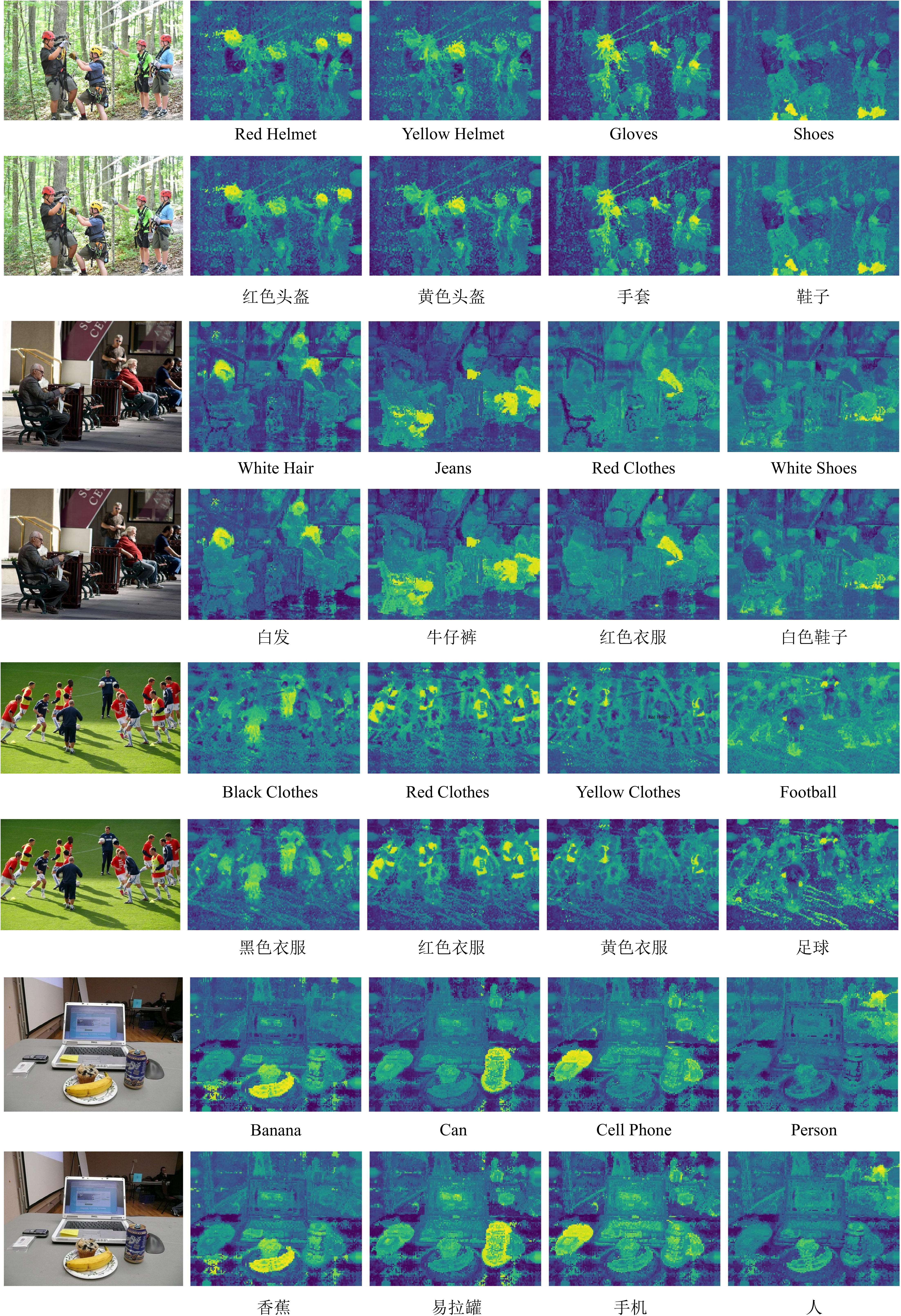}}
  
    \caption{Visualization of FG-CLIP 2's dense feature maps and semantic alignment capability in bilingual scenarios.}
    
    \label{fig:sup2}
\end{figure*}

We present the alignment capability of FG-CLIP 2 between dense visual features and text in both Chinese and English contexts. The results are shown in Figure~\ref{fig:sup2}, where warmer colors indicate higher similarity between image regions and the matched text. 
\iclrre{The precise localization of high-similarity regions across both languages demonstrates FG-CLIP~2’s strong bilingual semantic alignment and fine-grained perception capabilities.}

\section{\iclrre{Visual Analysis of the Textual Intra-modal Contrastive Loss}}
\label{appendix:tic}

\iclrre{We conduct an embedding visualization comparing text representations with and without the TIC loss. As shown in the visualization at Table~\ref{tab:tics1} and Table~\ref{tab:tics2}, when TIC is enabled, semantically similar but distinct phrases become better separated in the embedding space. This suggests that TIC indeed enhances semantic separability among fine-grained textual descriptions.}

\section{\iclrre{Examples of LIT-CN}}
\label{appendix:lit-cn}

\xie{In Table~\ref{tab:cn_longsample}, we provide examples of long caption image-text pairs from the LIT-CN dataset, covering diverse scene categories such as indoor, outdoor, animals, products, and buildings. These captions not only describe fine-grained subject attributes (e.g., appearance, posture, spatial layout), but also detail the surrounding context, reflecting the dataset’s semantic richness and descriptive complexity.}

\begin{table*}[!htbp]
\renewcommand\thetable{E.1}
  \centering
  \caption{\iclrre{The Effectiveness of the Textual Intra-modal Contrastive Loss. (Case1 and Case2)}}
  \resizebox{1.0\textwidth}{!}{
    \begin{tabular}{c}
    \toprule
    Case1\\
    \midrule
   \includegraphics[width=0.6\linewidth]{{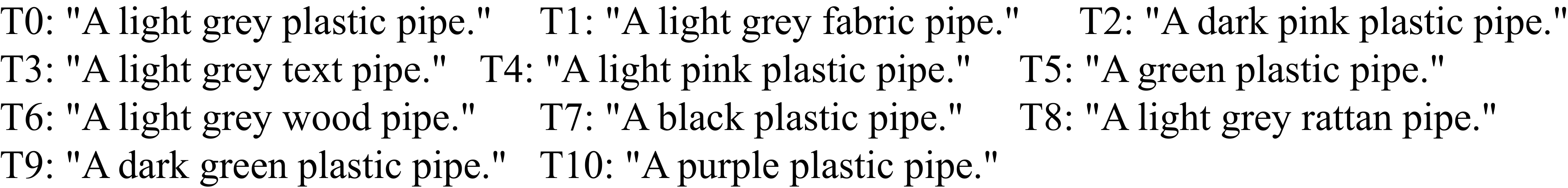}}\\
   \midrule
   \includegraphics[width=1.0\linewidth]{{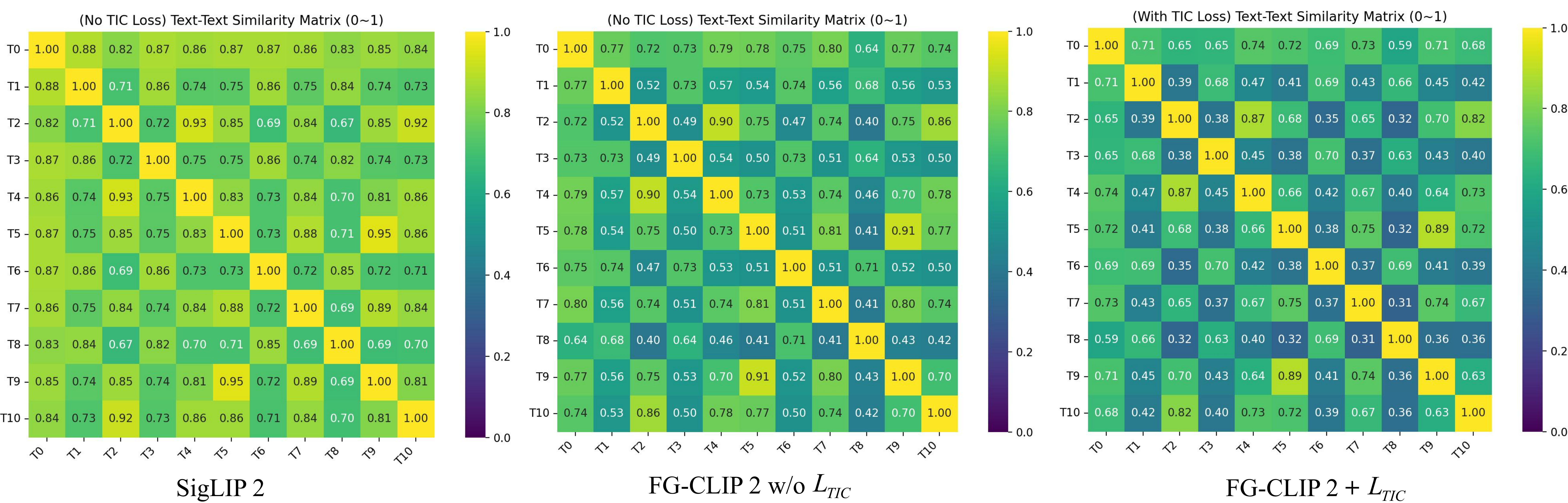}}\\
   \midrule
   Case2\\
    \midrule
   \includegraphics[width=0.85\linewidth]{{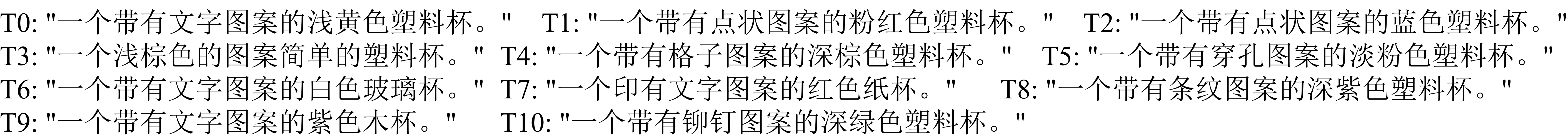}}\\
   \midrule
   \includegraphics[width=1.0\linewidth]{{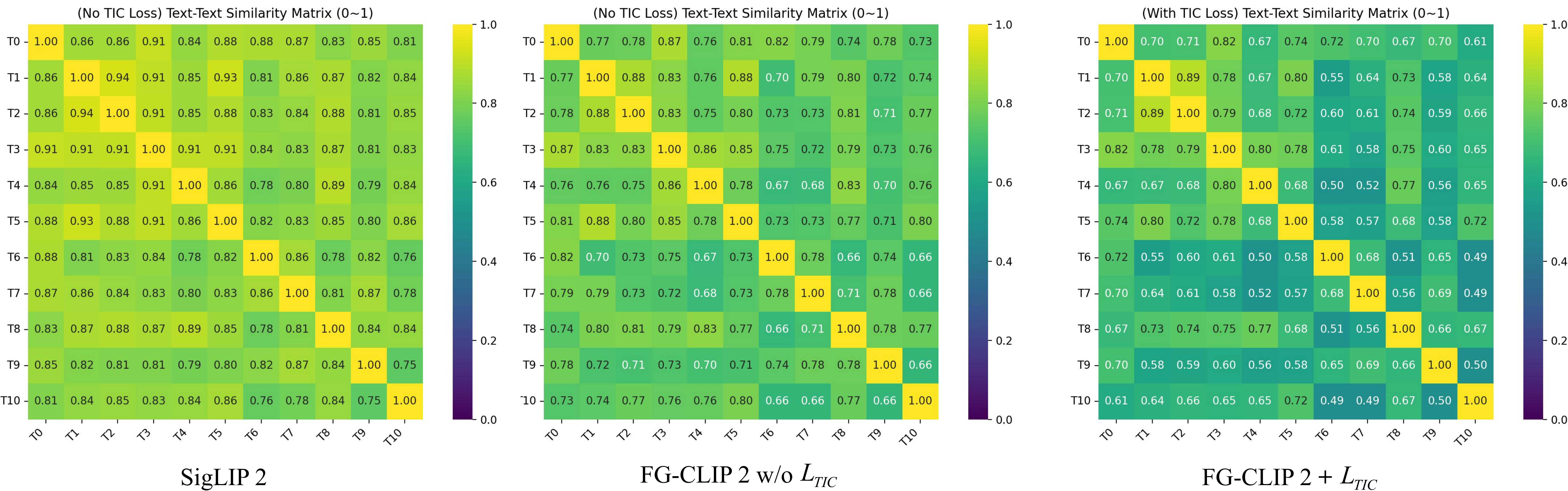}}\\
    \bottomrule
    \end{tabular}
  }
  \label{tab:tics1}
\end{table*}

\begin{table*}[!htbp]
\renewcommand\thetable{E.2}
  \centering
  \caption{\iclrre{The Effectiveness of the Textual Intra-modal Contrastive Loss. (Case3 and Case4)}}
  \resizebox{1.0\textwidth}{!}{
    \begin{tabular}{c}
    \toprule
   Case3\\
    \midrule
   \includegraphics[width=0.75\linewidth]{{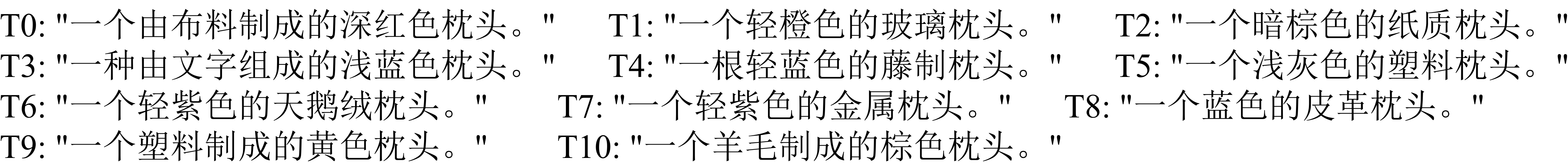}}\\
   \midrule
   \includegraphics[width=1.0\linewidth]{{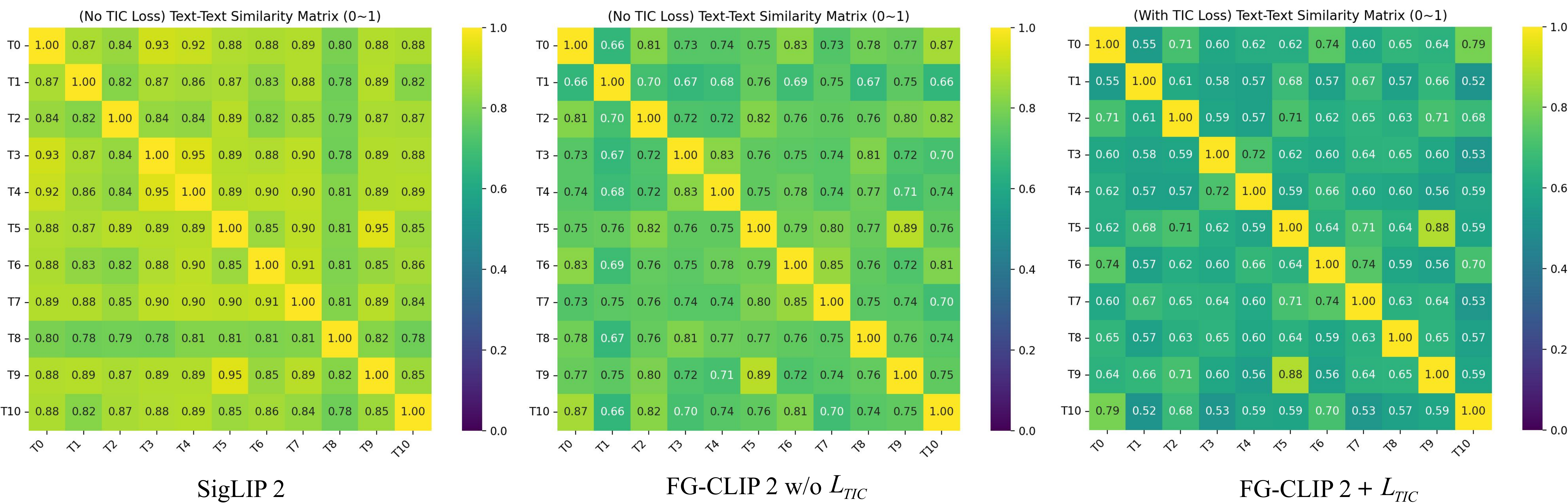}}\\
   \midrule
   Case4\\
    \midrule
   \includegraphics[width=1.0\linewidth]{{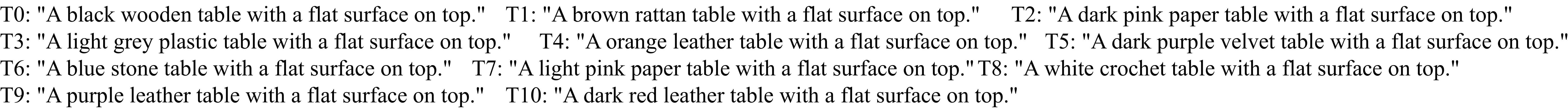}}\\
   \midrule
   \includegraphics[width=1.0\linewidth]{{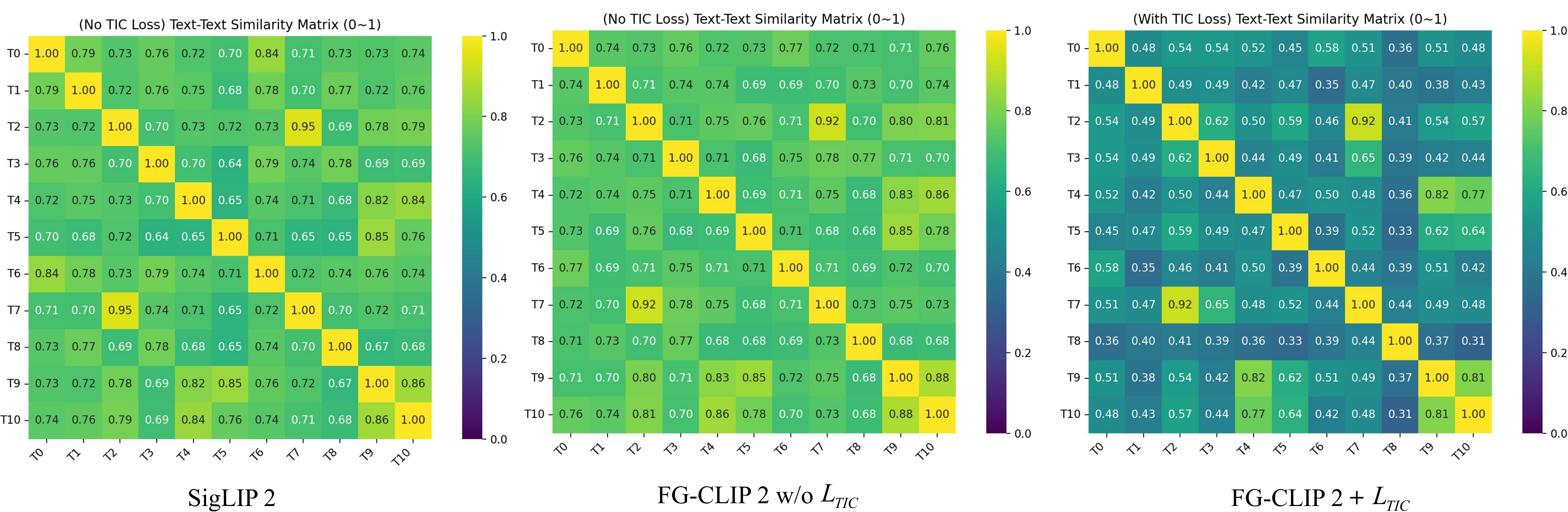}}\\
    \bottomrule
    \end{tabular}
  }
  \label{tab:tics2}
\end{table*}



\section{BoxClass-CN Category Schema and Example Explanation}
\label{appendix:boxclass-cn}
\begin{figure*}[!htbp]
    \renewcommand\thefigure{D}
  \centering   \includegraphics[width=1.0\linewidth]{{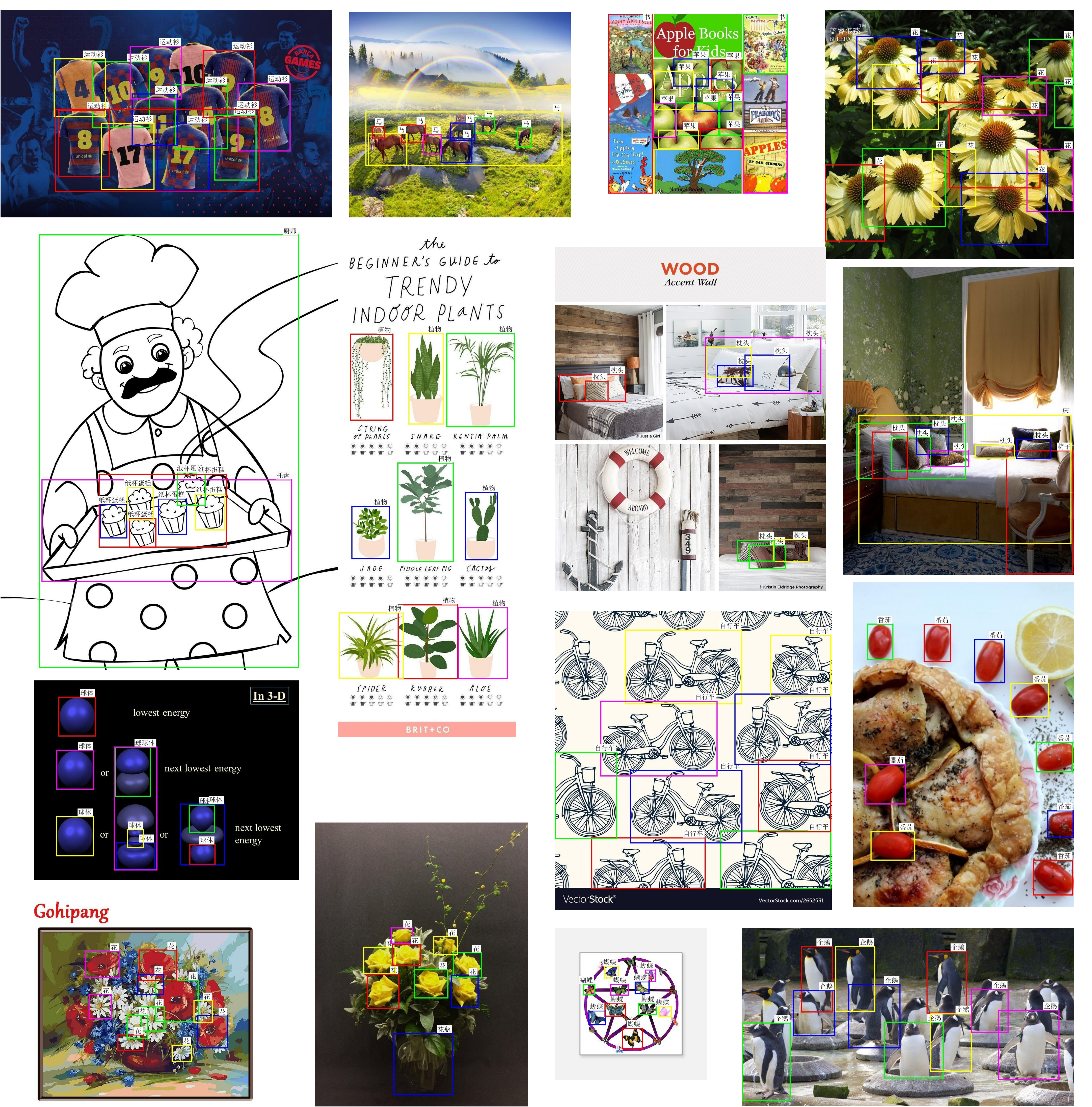}}
    \caption{Examples from BoxClass-CN.}
    
    \label{fig:sup1}
\end{figure*}

\xie{Table~\ref{tab:cn_cls1} and~\ref{tab:cn_cls2} provide the complete list of categories in the BoxClass-CN dataset, separated by commas. We further present some examples from BoxClass-CN in Figure~\ref{fig:sup1}.}




\begin{table}[!htbp]
\renewcommand\thetable{F}
  \centering
  \caption{Examples from LIT-CN.}
  \resizebox{0.85\textwidth}{!}{
    \begin{tabular}{cc}
    \toprule
\includegraphics[width=0.3\linewidth]{{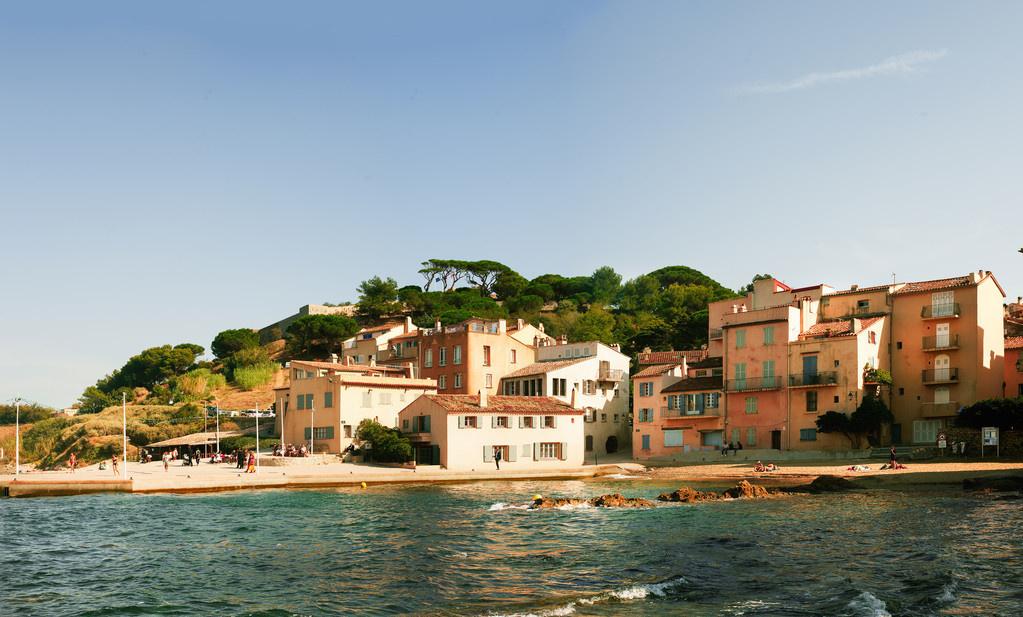}}&\includegraphics[width=1.0\linewidth]{{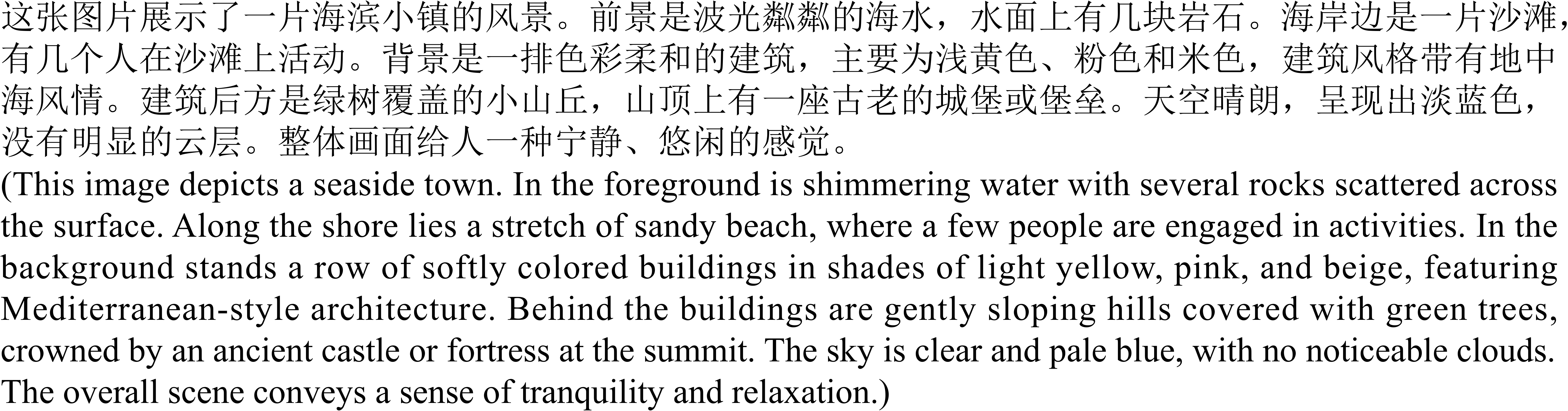}}\\
 \midrule
 \includegraphics[width=0.3\linewidth]{{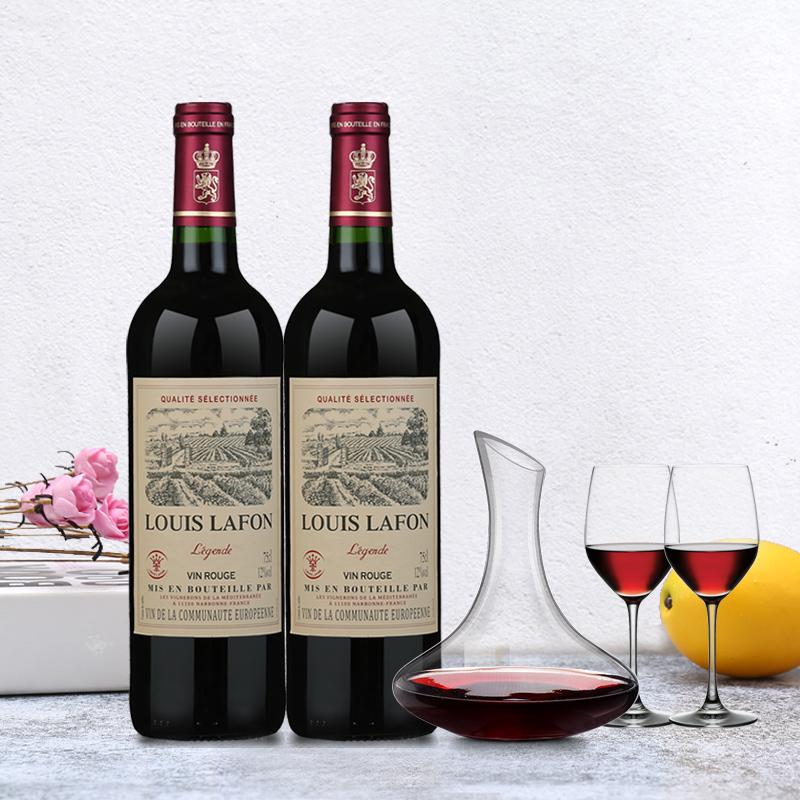}}&\includegraphics[width=1.0\linewidth]{{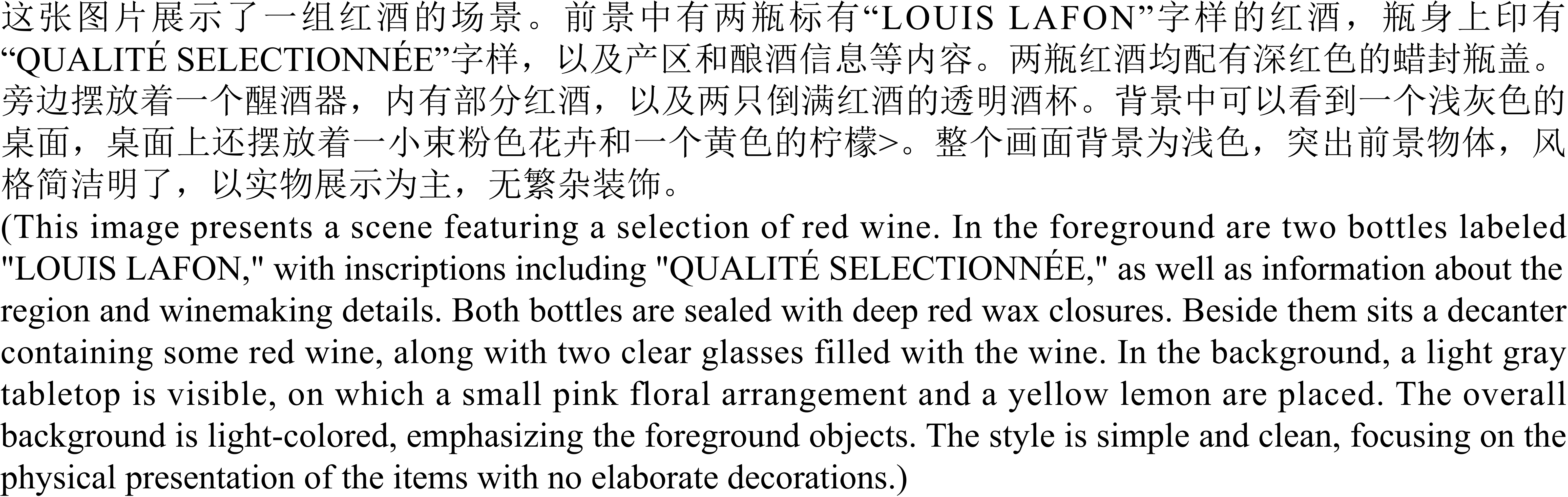}}\\
  \midrule
 \includegraphics[width=0.3\linewidth]{{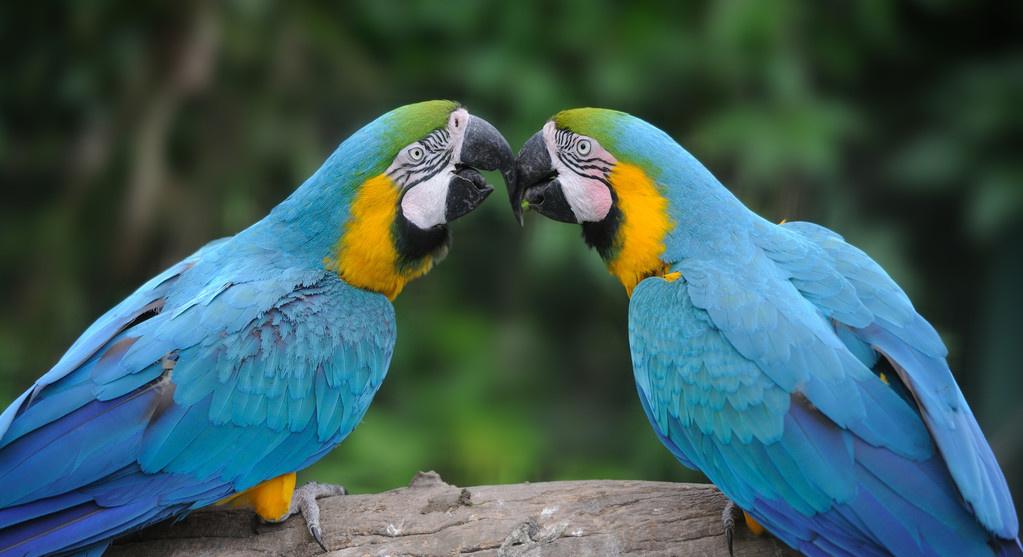}}&\includegraphics[width=1.0\linewidth]{{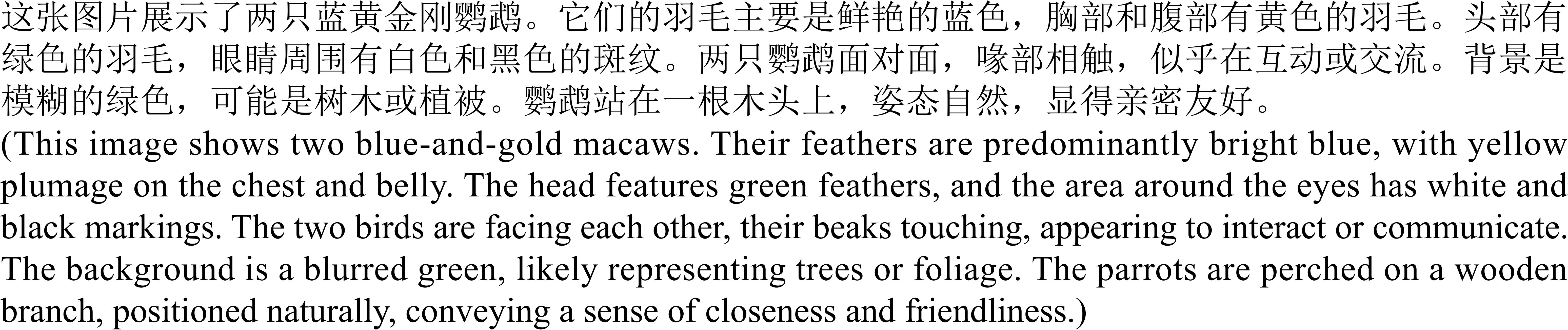}}\\
 \midrule
 \includegraphics[width=0.3\linewidth]{{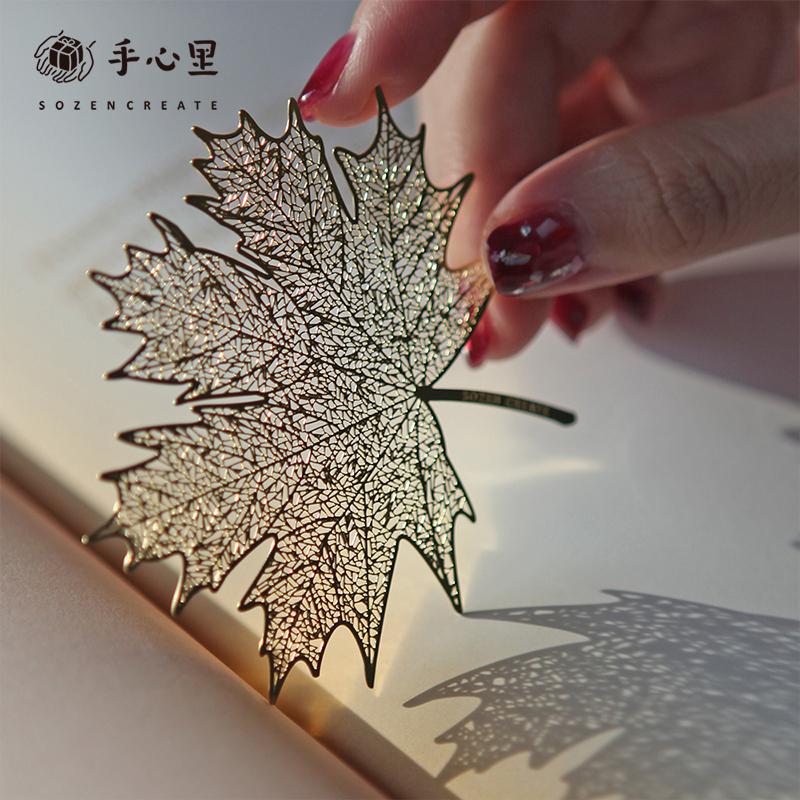}}&\includegraphics[width=1.0\linewidth]{{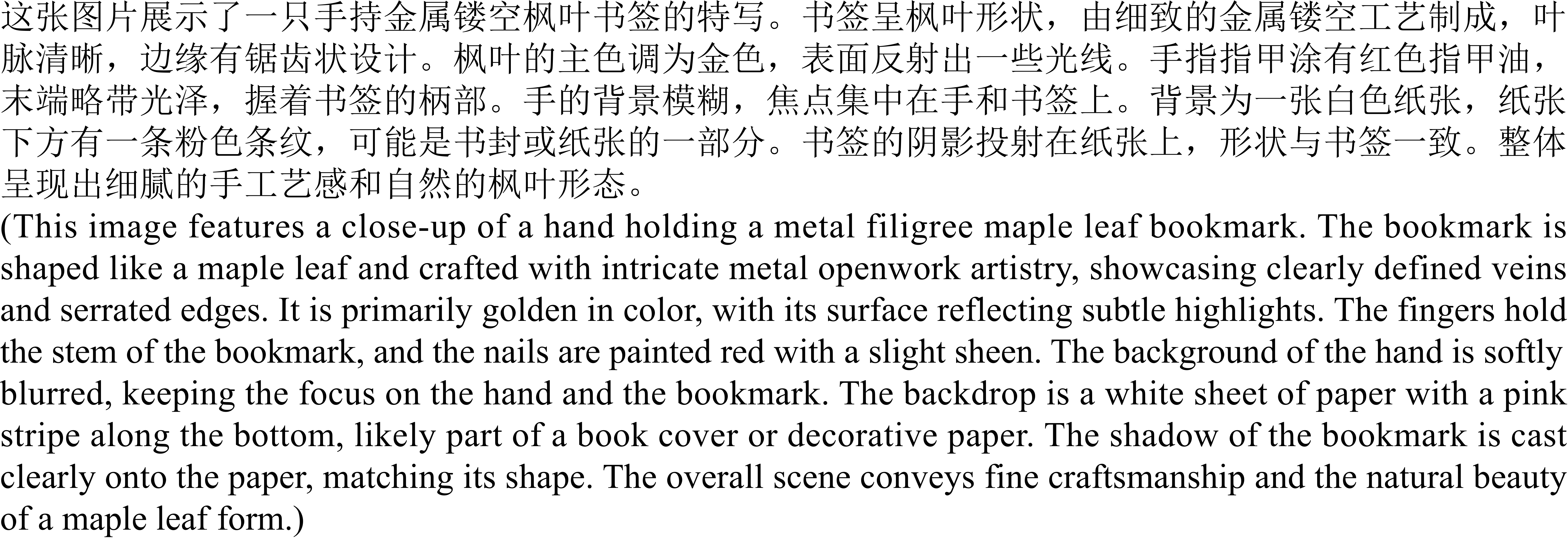}}\\
  \midrule
 \includegraphics[width=0.3\linewidth]{{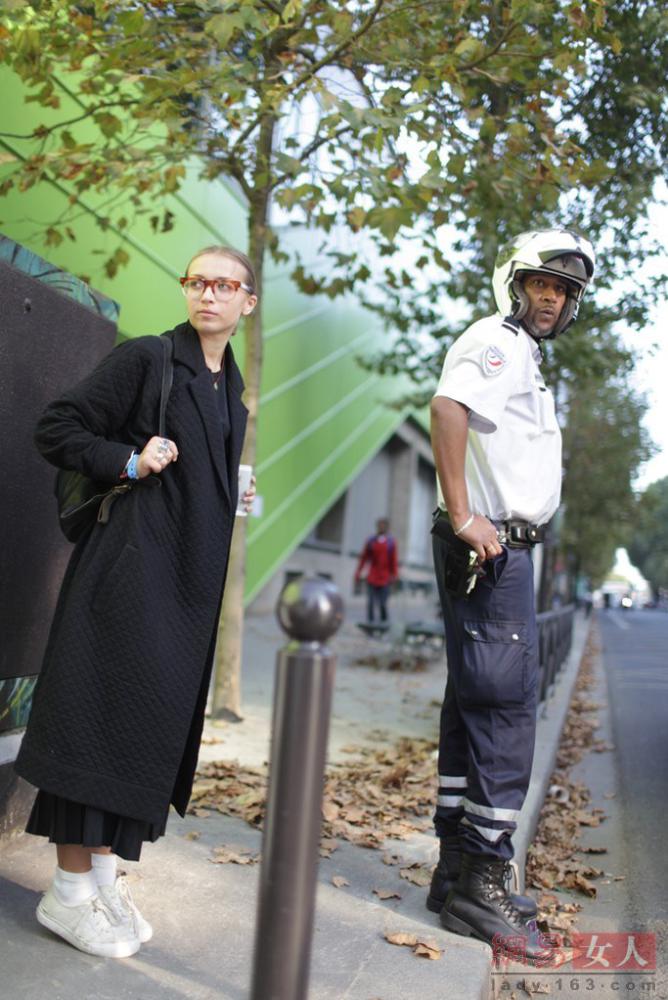}}&\includegraphics[width=1.0\linewidth]{{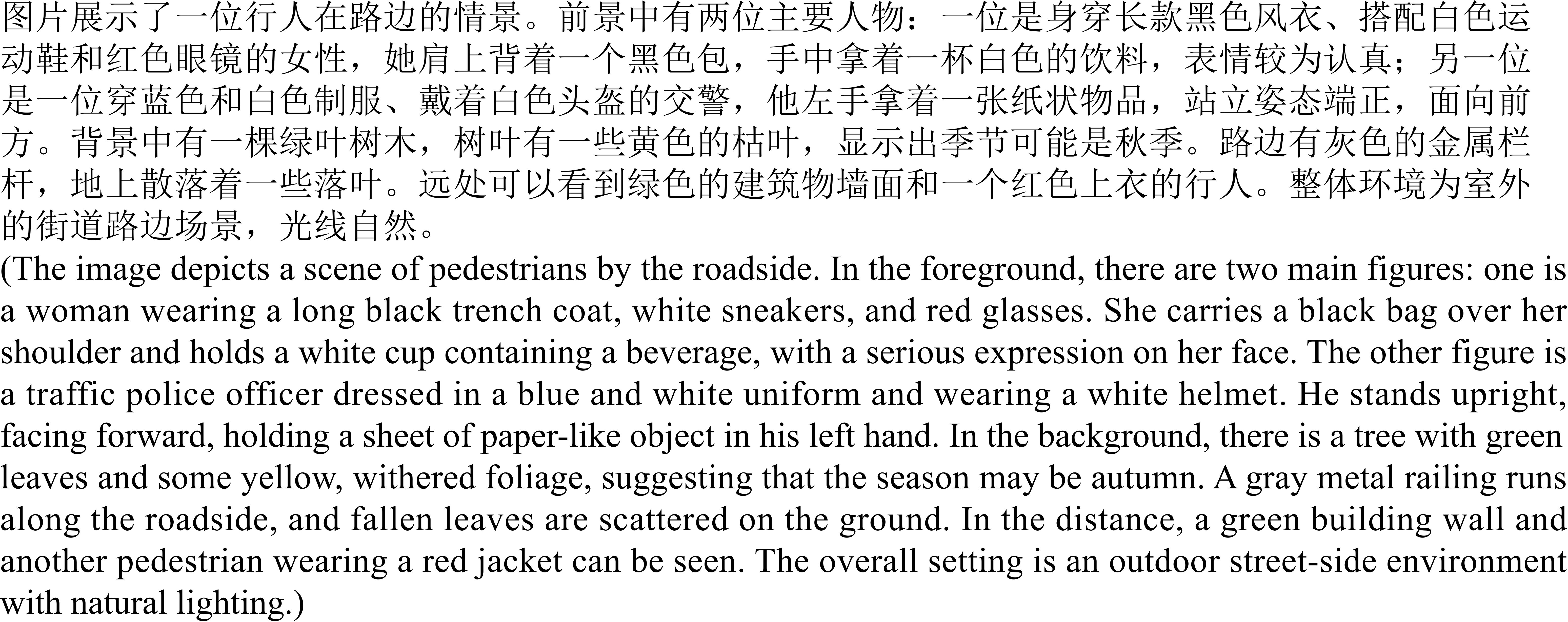}}\\
 \midrule
 \includegraphics[width=0.3\linewidth]{{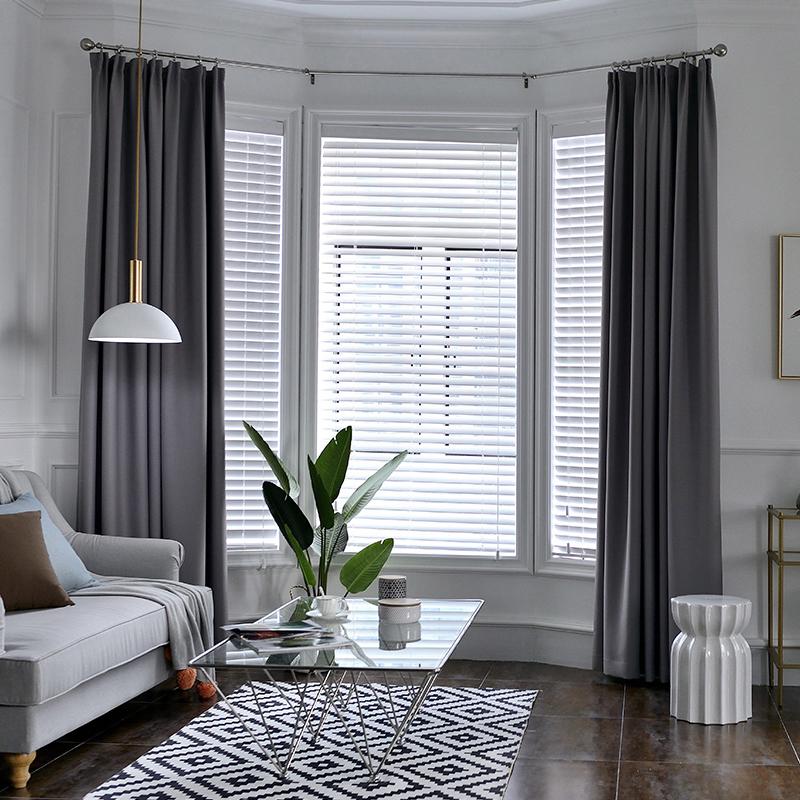}}&\includegraphics[width=1.0\linewidth]{{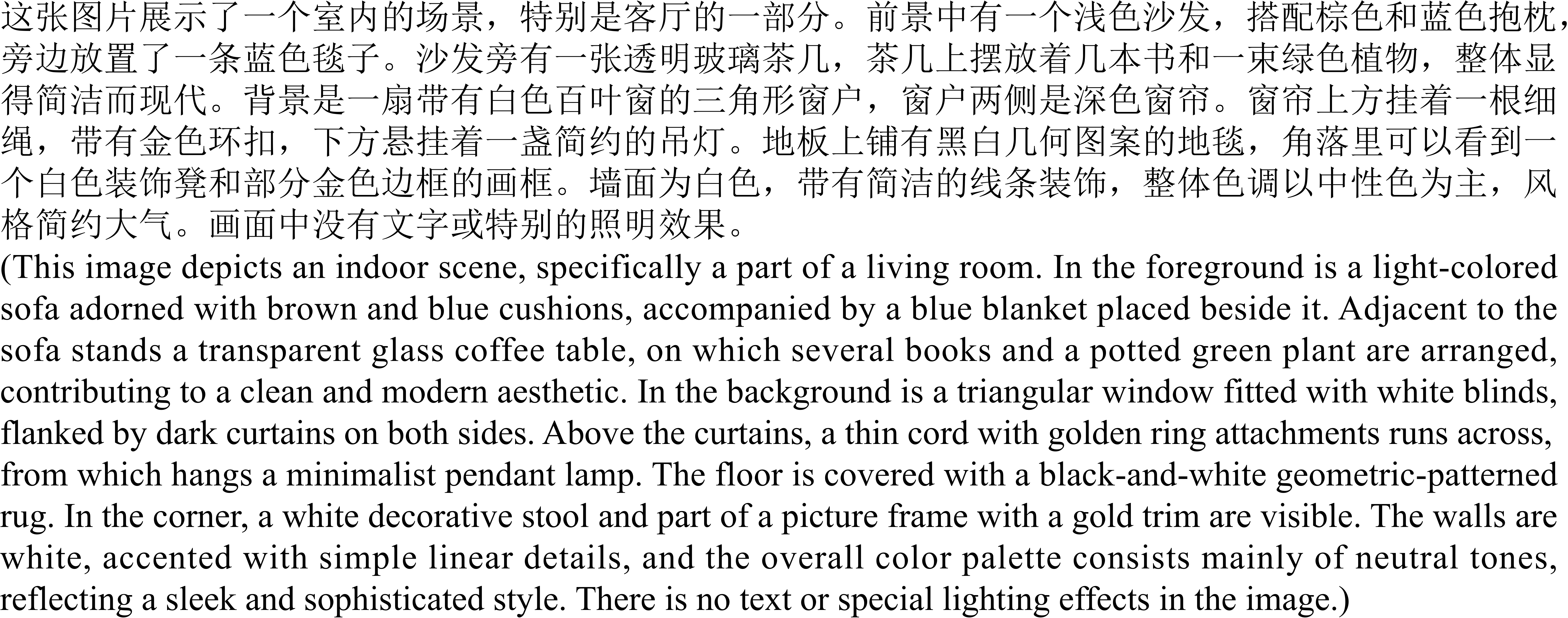}}\\
 \bottomrule
    \end{tabular}
  }
  \label{tab:cn_longsample}
\end{table}

\begin{table*}[h]
\renewcommand\thetable{G.1}
  \centering
  \caption{All categories in BoxClass-CN, grouped and displayed by ID (1-300).}
  \resizebox{1.0\textwidth}{!}{
    \begin{tabular}{c}
    \toprule
    id:1-100\\
    \midrule
   \includegraphics[width=1.0\linewidth]{{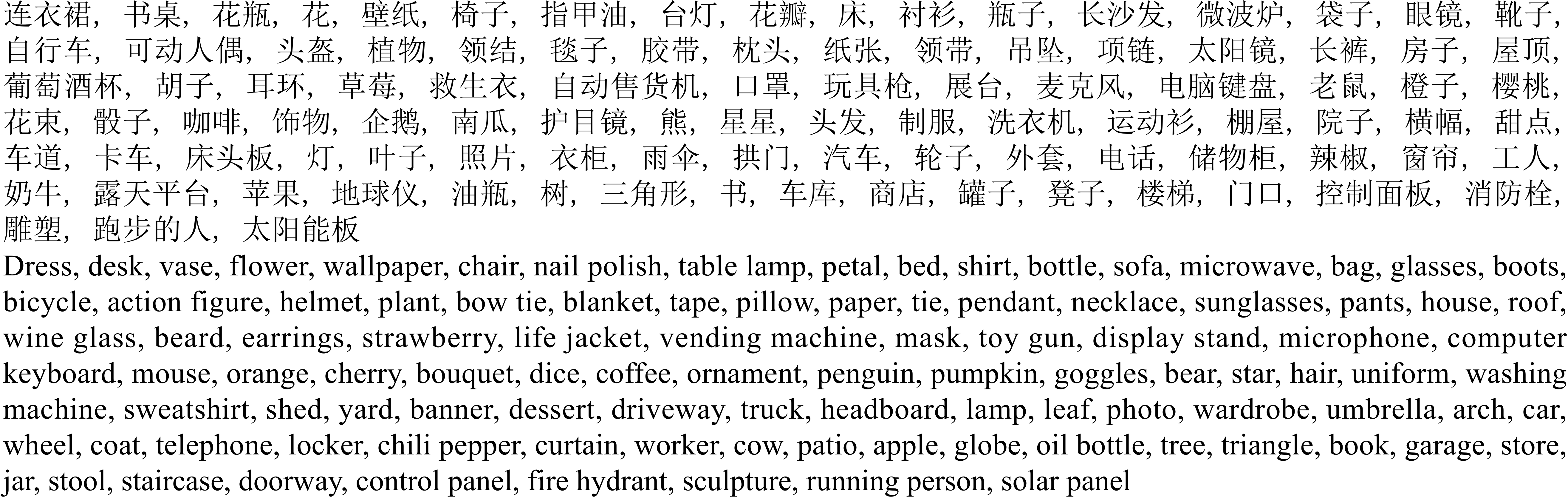}}\\
   \midrule
   id:101-200\\
   \midrule
   \includegraphics[width=1.0\linewidth]{{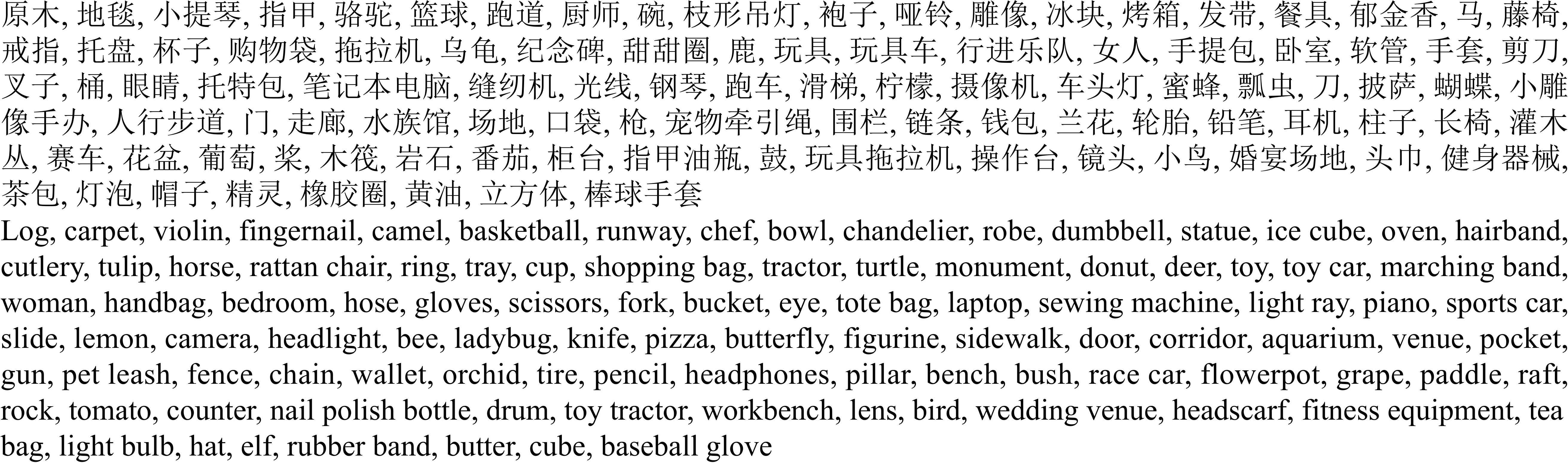}}\\
   \midrule
   id:201-300\\
   \midrule
   \includegraphics[width=1.0\linewidth]{{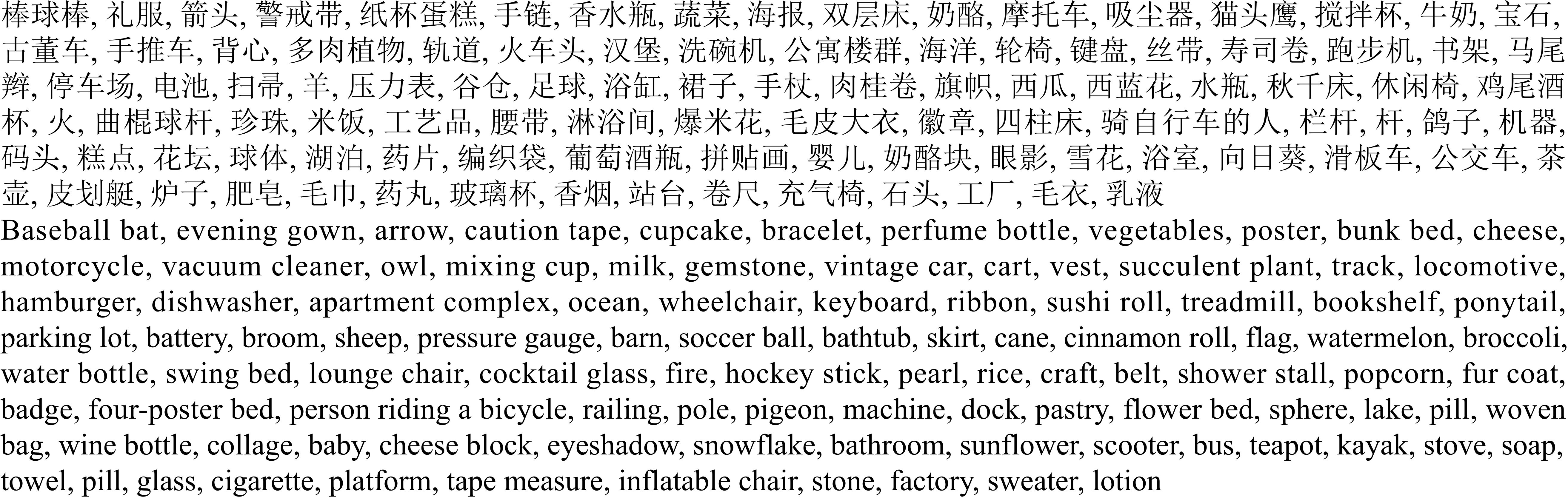}}\\
    \bottomrule
    \end{tabular}
  }
  \label{tab:cn_cls1}
\end{table*}

\begin{table*}[h]
\renewcommand\thetable{G.2}
  \centering
  \caption{All categories in BoxClass-CN, grouped and displayed by ID (301-566).}
  \resizebox{1.0\textwidth}{!}{
    \begin{tabular}{c}
    \toprule
   id:301-400\\
   \midrule
   \includegraphics[width=1.0\linewidth]{{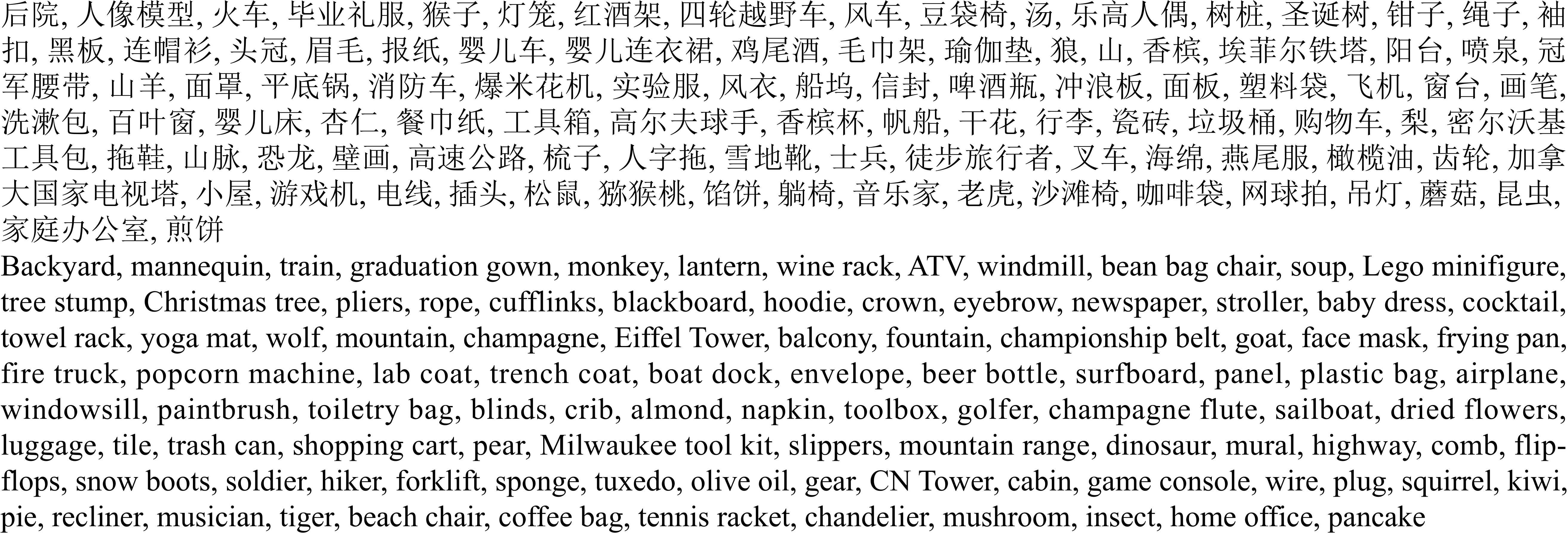}}\\
   \midrule
   id:401-500\\
   \midrule
   \includegraphics[width=1.0\linewidth]{{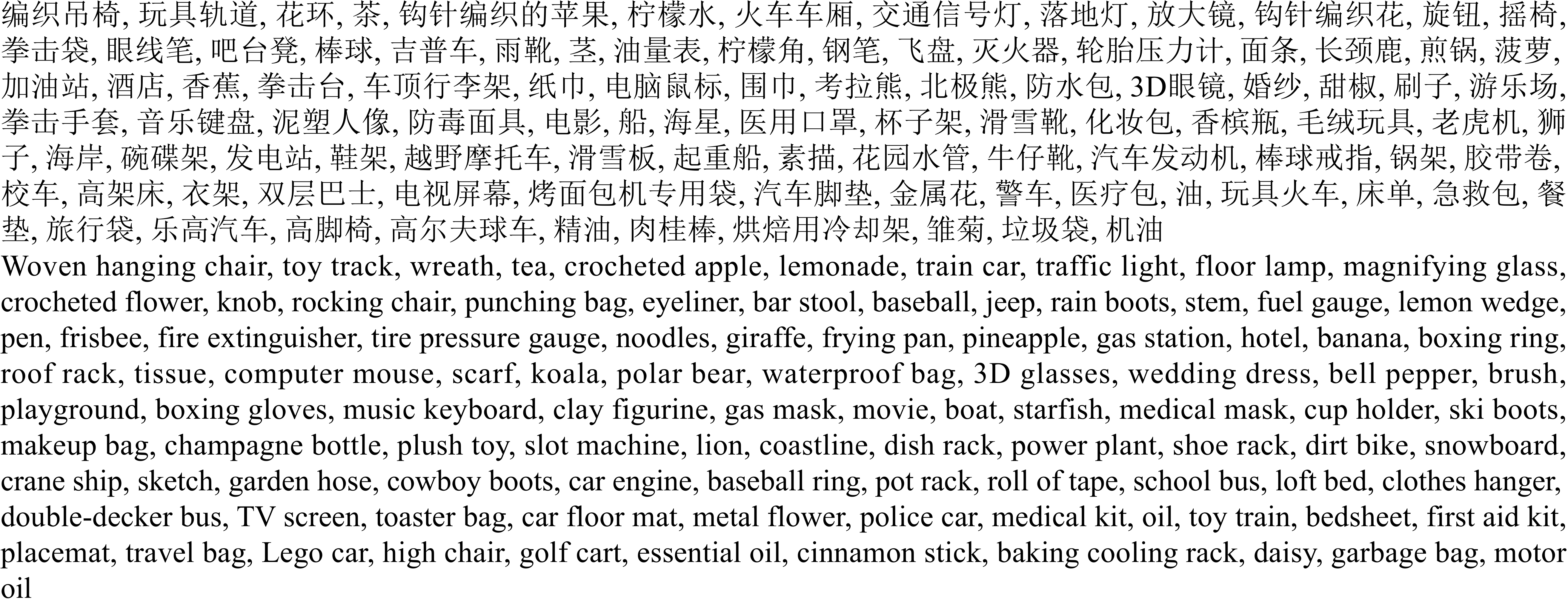}}\\
   \midrule
   id:501-566\\
   \midrule
   \includegraphics[width=1.0\linewidth]{{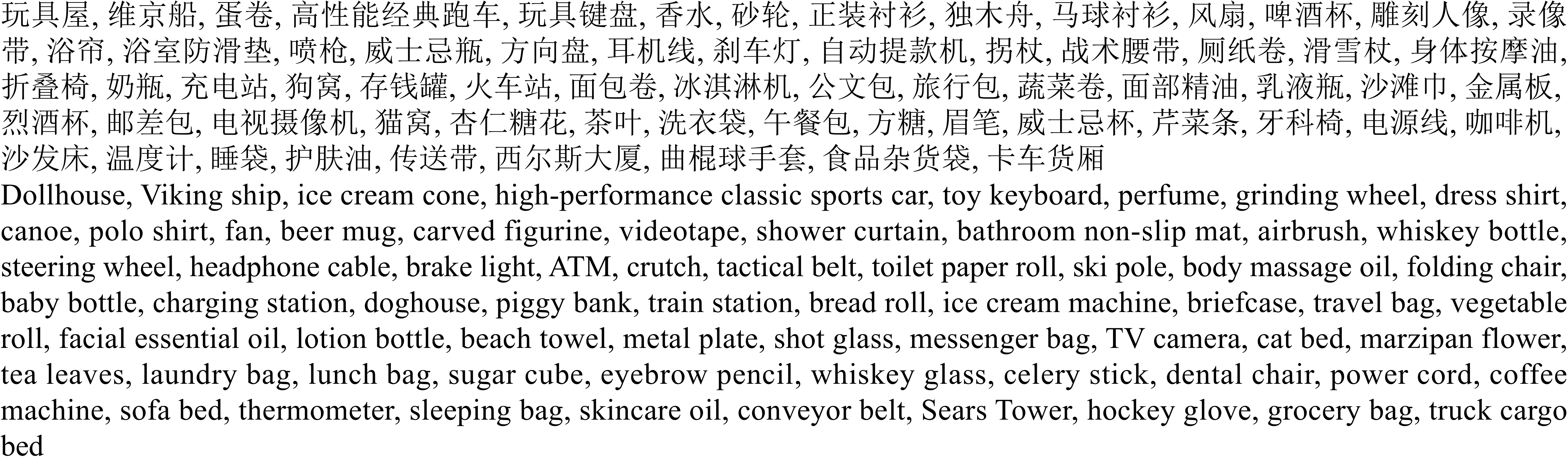}}\\
    \bottomrule
    \end{tabular}
  }
  \label{tab:cn_cls2}
\end{table*}

\end{document}